\documentclass[final,3p,times]{elsarticle}

\usepackage{amsmath}
\usepackage{lineno}
\usepackage{graphicx}
\usepackage{amsfonts}
\usepackage{amssymb}
\usepackage{hyperref}
\usepackage{xcolor}
\usepackage{comment}
\usepackage{booktabs}
\usepackage{multirow}
\usepackage{rotating}
\biboptions{sort&compress}

\journal{Journal of Network and Computer Applications}

\usepackage{fancyhdr}
\usepackage{url}
\fancypagestyle{firstpage}
{
    \fancyhead[L]{{\small{This is the author's version of the accepted manuscript in Journal of Network and Computer Applications. Formal publication can be found at \url{https://doi.org/10.1016/j.jnca.2025.104116}.© 2025. This manuscript version is made available under the CC-BY-NC-ND 4.0 license \url{https://creativecommons.org/licenses/by-nc-nd/4.0/}.} }}   
}

\begin{document}

\begin{frontmatter}




\title{A Review of Graph-Powered Data Quality Applications for IoT Monitoring Sensor Networks}


\author{Pau Ferrer-Cid}
\ead{pau.ferrer.cid@upc.edu}

\author{Jose M. Barcelo-Ordinas}
\author{Jorge Garcia-Vidal}

\affiliation{organization={Computer Architecture Department, Universitat Politècnica de Catalunya},
            city={Barcelona},
            country={Spain}}

\begin{abstract}
The development of Internet of Things (IoT) technologies has led to the widespread adoption of monitoring networks for a wide variety of applications, such as smart cities, environmental monitoring, and precision agriculture. A major research focus in recent years has been the development of graph-based techniques to improve the quality of data from sensor networks, a key aspect of the use of sensed data in decision-making processes, digital twins, and other applications. Emphasis has been placed on the development of machine learning (ML) and signal processing techniques over graphs, taking advantage of the benefits provided by the use of structured data through a graph topology. Many technologies such as graph signal processing (GSP) or the successful graph neural networks (GNNs) have been used for data quality enhancement tasks. This survey focuses on graph-based models for data quality control in monitoring sensor networks. In addition, it introduces the technical details that are commonly used to provide powerful graph-based solutions for data quality tasks in sensor networks, such as missing value imputation, outlier detection, or virtual sensing. To conclude, different challenges and emerging trends have been identified, e.g., graph-based models for digital twins or model transferability and generalization.
\end{abstract}



\begin{keyword}
Data Quality \sep Graph Signal Processing \sep Graph Neural Networks \sep Internet of Things \sep Machine Learning \sep Monitoring Sensor Networks


\end{keyword}

\end{frontmatter}

\thispagestyle{firstpage}
\section{Introduction}
\label{sec:introduction}

The GSP field emerged to translate the classic signal processing concepts, related to the analysis of time-dependent signals, to irregular domains such as graphs, defining the notion of signals defined over graphs \cite{shuman2013emerging, sandryhaila2013discrete}. The use of graph-based models had already been a key element in applications such as route planning (e.g., Dijkstra's algorithm) \cite{delling2009engineering}, community detection (e.g., clique percolation or Louvain algorithms) \cite{fortunato2010community} or the analysis of complex networks such as biological and social networks \cite{pavlopoulos2011using, newman2002random}. The representation of manifolds by means of graphs in the field of semi-supervised learning is another example of graph-powered applications \cite{belkin2004semi}. Overall, the GSP framework \cite{leus2023graph} has enabled the use and development of novel techniques on data residing in graphs, thus emerging as an alternative to classical ML techniques that do not make explicit use of data structure. In this way, the graph topology, which represents the relationships between the graph's nodes, is fed to graph-based models that explicitly model the structure of the data \cite{dong2020graph}.

A wide variety of concepts have been applied to signals defined over graphs, such as signal shift, translation, convolution, or filtering \cite{ortega2018graph}. An important concept of the GSP is the notion of signal smoothness, also expressed via the total variation (TV) or the Dirichlet energy, which are quadratic forms and can be used to evaluate how a signal fits a given graph structure or vice-versa \cite{zhang2015graph}. This idea is linked with the Graph Discrete Fourier Transform (GDFT) that makes use of the graph topology to obtain the graph Fourier basis and allow the computation of the transform coefficients of a graph signal and also led to the development of graph filters \cite{sandryhaila2013discretefourier}. In the field of ML, graphs have been used as regularizers in optimization problems, e.g., the regularization of neural networks for semi-supervised learning tasks \cite{chami2022machine}. Moreover, the structured representation of the data allows for improving the interpretability of the models as well as the interpretability of their results \cite{dong2020graph}. Given the growing popularity of black-box models (e.g., deep neural networks), much emphasis has been placed on the interpretability of the techniques to facilitate their use. In addition, one of the greatest successes of graph-based techniques has been the development of GNNs, allowing the use of neural networks on data residing in irregular domains such as graphs, from more general message-passing-based GNNs to spectral networks based on the graph convolution operation \cite{zhou2020graph, wu2022graph}.

\begin{figure}[!h]
    \centering
    \includegraphics[width=0.75\textwidth]{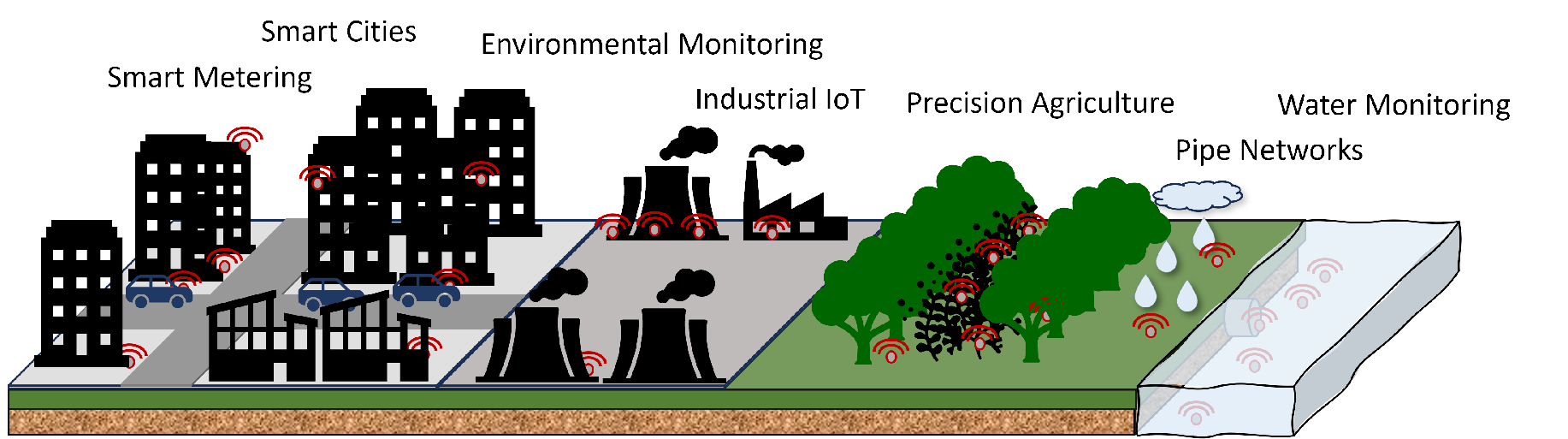}
    \caption{Examples of different applications where IoT sensor networks are leveraged to monitor different phenomena. Recent use cases include applications in air quality networks, precision agriculture, or pipe networks among others.}
    \label{fig:monitoring_appls}
\end{figure}

Recently, graphs have unveiled their potential to represent IoT monitoring sensor networks, allowing the use of signal processing and ML techniques to carry out applications using sensor network data \cite{jablonski2017graph, dong2023graph}. Monitoring sensor networks are platforms whose purpose is to monitor the evolution of certain phenomena in a specific area of interest or scenario, e.g., precision agriculture systems \cite{lopez2022low}, air quality monitoring \cite{ferrer2020multisensor}, or industrial processes \cite{wu2021graph, su2024graph}. These networks are of particular interest because they monitor critical phenomena or assets and provide data for further applications or decision-making processes. Figure \ref{fig:monitoring_appls} depicts different scenarios where IoT monitoring networks are used, common use cases include smart metering, industrial applications, and environmental monitoring among others.

Data-driven IoT applications aim to exploit the network measurements as well as their intrinsic structure to carry out several tasks. The quality of data from monitoring sensor networks is critical since the data can feed various applications or decision-making processes that require complete and accurate data. In the literature, data quality tasks have been developed to provide continuous, complete, and accurate data. Data quality in IoT platforms is defined by different dimensions such as data accuracy, sample completeness, data consistency or coherence, and data timeliness \cite{guerra2023iso, mansouri2023iot,liu2020data}. Some of the issues that may affect such data attributes are the presence of missing data, outliers, inconsistent samples, etc. Therefore, some of the tasks developed include the detection of outliers and anomalies \cite{ferrer2022volterra}, the use of predictive models to create virtual sensors \cite{yan2022virtual}, the imputation of missing data \cite{jiang2021graph}, and the detection of clusters and communities among others \cite{jablonski2017graph}. In the context of sensor data, models have also been developed to estimate the reliability of the measurements, an aspect related to possible measurement deviations and thus outlying or anomalous measurements \cite{shafin2024sensor}. Indeed, graph-based models provide alternatives to classic data-driven approaches and have been proven to be superior in terms of performance, possibility of distributed approaches, and interpretability. IoT data is characterized by the presence of spatiotemporal correlations given that monitoring platforms measure phenomena in an area of interest or assets in an industrial environment. In particular, graph-based models are known for their ability to model complex relationships between sensors and leverage spatiotemporal correlations \cite{dong2023graph}. Henceforth, a deep understanding of signal processing and ML over graph approaches may unveil new applications and advantages for IoT monitoring sensor networks.

Recent surveys are based on the technical review of models based mainly on GSP and GNN, mentioning some applications in areas such as computer vision or natural language processing \cite{zhou2020graph,song2022graph, wu2020comprehensive, jin2023survey, xia2021graph}. Others focus on the applications of GNNs and graph-based models in areas such as the IoT, showing graph-based applications in cases such as environmental monitoring or biological data \cite{dong2023graph,li2023graph,li2021graphs, wang2023artificial}.
In this survey, we focus on the application of graph-based models to the recently booming area of data quality improvement for monitoring IoT platforms \cite{ferrer2024data, wang2023artificial}. In particular, we not only cover GNNs applications but also provide complete coverage of graph-based modeling of monitoring networks, covering GSP, ML over graphs, and GNNs. We also present recent advances and applications in data quality control and provide a taxonomy for the most novel graph-based applications and trends.

\begin{figure*}[t]
    \centering
    \includegraphics[width=0.925\textwidth]{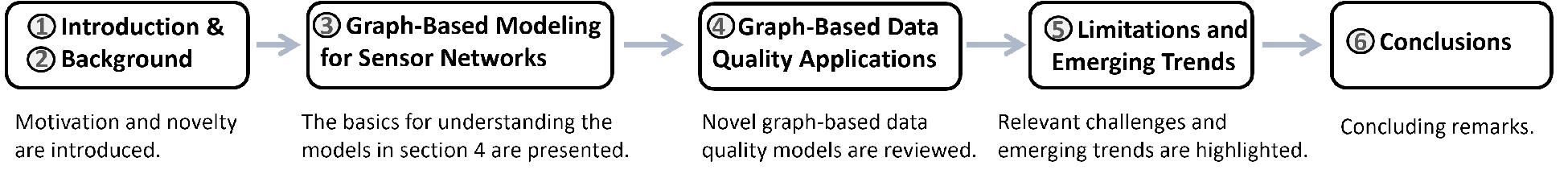}
    \caption{Outline of the survey. Section \ref{sec:technical_part}  introduces the basics of the different graph-based approaches (GSP, ML over graphs, and GNNs) on which the methods discussed in section \ref{sec:applications} are based.}
    \label{fig:outline}
\end{figure*}

The outline of this paper is as follows; Section \ref{sec:background} motivates the content of the paper. Section \ref{sec:technical_part} introduces technical concepts regarding signal processing and ML over graphs. Section \ref{sec:applications} reviews different graph applications in IoT monitoring sensor networks, and section \ref{sec:future} presents challenges, limitations, and emerging trends. Finally, section \ref{sec:conclusions} concludes the manuscript. Figure \ref{fig:outline} shows the outline of the survey.

\section{Background and Motivation}
\label{sec:background}
In this section, we revisit the development of graph-based techniques and we motivate the use of graph-based models for IoT data quality applications.

Before the appearance of the GSP, graphs were already used in the field of graph-based semi-supervised learning to present the relationships of data residing in a manifold \cite{song2022graph, subramanya2022graph}. In fact, label propagation emerged as a transductive method to estimate unlabeled data points from labeled ones, and inspired the GNN framework \cite{wang2020unifying}. The advent of the GSP was a revolution since it allowed the translation of solid signal processing fundamentals to the realm of signals defined over graphs. Thus, this allowed the analysis of measurements residing on a graph via graph filters, signal shifts, signal convolutions, or the spectral graph Fourier transform among others. In turn, the graph signal convolution led to the translation of convolutional neural networks to the graph paradigm, making use of the graph signal convolution operator, e.g., the Chebnet GNN \cite{defferrard2016convolutional}. Besides all the guarantees in terms of efficiency and accuracy, the GSP-based techniques provide interpretability, which has received a lot of attention due to the growing use of black-box models, given the use of the graph shift operator, i.e., the matrix that describes the relationships between network nodes. Moreover, many of the techniques, e.g., graph filters, allow a distributed implementation, which is very interesting in case of resource limitations or in case of having nodes with computational resources, e.g., edge computing approach \cite{coutino2019advances}. All this together makes the GSP provide appropriate tools to analyze the measurements produced by a network of sensors, being able to define these measurements as signals that reside in a graph.

ML over graphs usually refers to ML techniques that have been adapted to take into account that a graph can represent the data, or analogously that the data resides in a manifold. In this sense, graph-based semi-supervised learning is aimed at the application of algorithms to partially-labeled data residing on manifolds represented by graphs \cite{belkin2004semi}. This idea has also been used in different works where the optimization function of different ML techniques has been modified to include a regularization term that takes into account that the data points reside on a graph \cite{hang2016graph}. Among some examples, we can find predictive models, e.g., kernel ridge regression \cite{venkitaraman2019predicting} or regularized support vector regression \cite{li2020semi}, which can be very useful in the case of sensor network measurements. Even though, there exists a large range of tasks and applications that can be approached using ML, e.g., the target of the ML task can be estimating features at graph nodes, at graph edges, or at the graph as a whole. Throughout the manuscript, we restrict to the tasks that are useful in the sensor network domain.

Although GNNs belong to the field of ML over graphs, in this paper we want to dedicate their own space to them due to their great importance and widespread adoption nowadays. Some well-known architectures include the GraphSAGE \cite{hamilton2017inductive} or the Chebnet \cite{defferrard2016convolutional}. Indeed, the graph convolution operator allowed the application of the convolution neural networks to the graph realm, resulting in graph convolutional neural networks such as the Chebnet, in which the convolution is realized via a graph filter. Given the widespread application of neural networks, tailored GNN architectures are being continuously developed to tackle tasks such as missing value imputation, anomaly detection, data compression, event detection, or signal reconstruction \cite{zhou2020graph}. Another example of disruption in this field is the emergence of graph attention networks and graph transformers, being the analogs of attention mechanisms and transformers in conventional neural networks \cite{ahmad2021gate}.

Recently, graphs have been used to model measurements from IoT platforms in industrial, smart city, or precision agriculture applications among others \cite{dong2023graph}. In this way, structured data modeling using graphs has been exploited, either by GSP or GNN \cite{ortega2018graph,egilmez2014spectral, wu2023detecting}. Some of the applications include structure learning, multivariate time series forecasting, and event or anomaly detection \cite{CALO2024108191,chen2021learning, xiao2020anomalous}. Recent studies have investigated the use of GNN in IoT applications \cite{dong2023graph, li2023graph, wang2023artificial}. Nevertheless, although there are many methods intended for data quality tasks, there is no survey that covers all these methods and frameworks in the context of data quality improvement in IoT monitoring networks.

\begin{figure*}[t]
    \centering
    \includegraphics[width=0.90\textwidth]{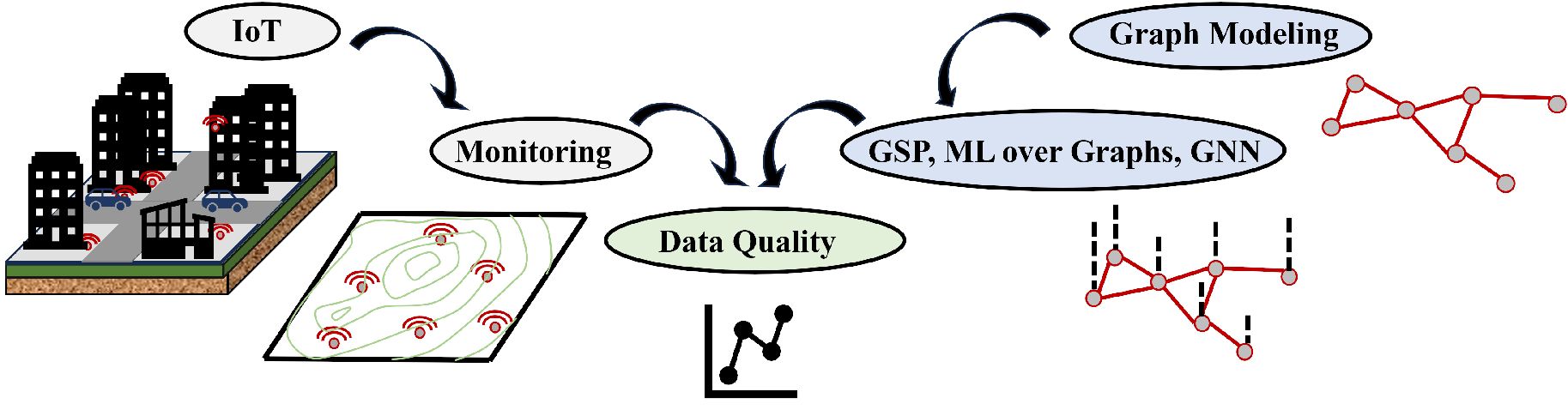}
    \caption{Scope of the review, focus on graph-based models (GSP, ML over graphs, and GNNs) for data quality tasks in IoT monitoring sensor networks.}
    \label{fig:scope_v2}
\end{figure*}

Figure \ref{fig:scope_v2} depicts the scope of this review. From now on we focus on GSP, ML over graphs, and GNNs, placing special emphasis on novel applications for data quality in monitoring sensor networks. Data quality is a key challenge in monitoring sensor networks, with tasks ranging from anomaly detection to missing value imputation \cite{ferrer2024data}. Although existing surveys cover various topics related to GSP and GNN for IoT (Table \ref{tab:novelty}), the review of graph-based models for data quality in IoT has not yet been covered. In addition, the survey groups into useful applications (in the context of IoT data quality) different works that have developed methods for different applications in isolation. Hence, this survey provides a novel description of graph-based models for data quality tasks in IoT monitoring sensor networks. In summary, our contributions are:
\begin{itemize}
    \item describe the foundations of different graph-based models for IoT sensor networks (GSP, ML over graphs, and GNNs) that can be used in data quality enhancement applications,
    \item review recent work and provide a taxonomy of applications that can help practitioners implement graph-based models in monitoring sensor networks to improve data quality and ensure data quality requirements,
    \item highlight the challenges and limitations that practitioners may face, as well as identify emerging trends.
\end{itemize}

\begin{table}[!htp]
\centering
\caption{A comparison between the scope of this survey and the scope of other relevant surveys.}
\label{tab:novelty}
\resizebox{0.75\columnwidth}{!}{
\begin{tabular}{@{}cccccc@{}}
\toprule
         & \textbf{GSP}  & \textbf{ML Over Graphs} & \textbf{GNN}  & \textbf{IoT Monitoring}   & \textbf{Data Quality} \\ \midrule
         \cite{teh2020sensor, mansouri2023iot} & & & & \checkmark & \checkmark \\
\cite{leus2023graph,ortega2018graph, dong2019learning, xia2021graph}      & \checkmark &                &      &                  &              \\
\cite{chami2022machine,song2022graph}      &  &     \checkmark           &      &                 &              \\
\cite{wu2020comprehensive, wu2022graph, zhou2020graph, zhang2019graph, jin2023survey}
&  &                &   \checkmark   &                 &              \\

\cite{dong2020graph,li2021graphs}      & \checkmark &     \checkmark           &   \checkmark   &                  &              \\
\cite{dong2023graph,li2023graph,wang2023artificial}
&  &                &   \checkmark   &     \checkmark      &              \\
Our Work & \checkmark & \checkmark           & \checkmark & \checkmark       & \checkmark         \\ \bottomrule
\end{tabular}}
\end{table}

\section{Graph-Based Modeling for Sensor Networks}
\label{sec:technical_part}
In this section, we elaborate on the fundamentals of the GSP, ML over graphs, and GNNs, highlighting current trends and novel techniques. We denote the measurement of a given sensor $i$ at a given time $t$ as $x_i(t){\in}\mathbb{R}$, $N$ denotes the number of sensors, and $\mathbf{x}_i{\in}\mathbb{R}^T$ represents a vector of $T$ measurements recorded by the $i$-th sensor. Matrices are denoted by bold uppercase letters, vectors by bold lowercase letters, and sets by calligraphic letters.

\begin{figure*}[!ht]
    \centering
    \includegraphics[width=0.4\columnwidth]{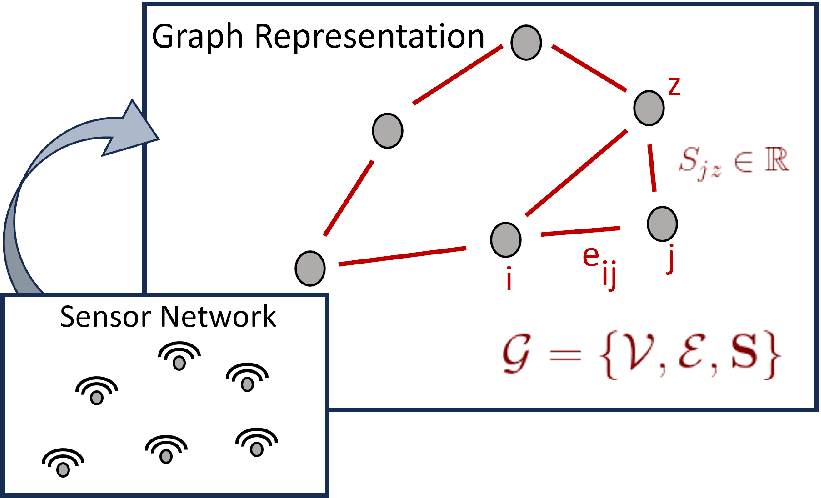}
    \caption{Example of how a graph $\mathcal{G}=\{\mathcal{V},\mathcal{E},\mathbf{S}\}$ can be used to represent a sensor network and the sensors' relationships. $e_{ij}$ represents an edge connecting nodes $i$ and $j$. The entry of the graph shift matrix $S_{ij}$ assigns a weight to the edge $e_{ij}$.}
    \label{fig:sensor_network}
\end{figure*}

\textit{Sensor network graph representation:} a monitoring sensor network and the relationship between nodes' measurements can be represented as a graph $\mathcal{G}$. In turn, a graph can be fully described by the triplet $\mathcal{G}{=}\{\mathcal{V}, \mathcal{E}, \mathbf{S}\}$, where $\mathcal{V}{=}\{1, \dots, N\}$ is the set of nodes representing the sensors, $\mathcal{E}{\subseteq} \{(i, j){:\;} i,j{\in}\mathcal{V}\}$ denotes the set of edges representing connected nodes, i.e., related sensors, and $\mathbf{S}{\in}\mathbb{R}^{N{\times} N}$ is a graph matrix describing the edges $\mathcal{E}$ of the graph, i.e., $\mathbf{S}_{ij}{\neq} 0$ iff $(i,j){\in}\mathcal{E}$. Figure \ref{fig:sensor_network} represents a sensor network described by graph $\mathcal{G}$. Graph structure identification, i.e., constructing the graph representing a sensor network is a research field on its own and it is further explained in section \ref{subsubsec:shift}. While a graph can denote \textit{spatial}, \textit{temporal}, and \textit{spatiotemporal} dependencies between nodes' data, spatiotemporal graphs can be constructed by combining a spatial graph $\mathcal{G}_s$ and a temporal graph $\mathcal{G}_t$ via their cartesian product $\mathcal{G}_{st}{=}\mathcal{G}_s{\times} \mathcal{G}_t$\footnote{There exist several ways to combine the structure of two graphs, e.g., cartesian product or Kronecker product. Refer to \cite{stankovic2020datapart1} for further information on graph combination.}. Another possible data-driven approach for learning spatiotemporal graphs consists of rearranging the sensors' measurements so that time lags are included as features.

\begin{figure*}[!ht]
    \centering
    \includegraphics[width=0.55\columnwidth]{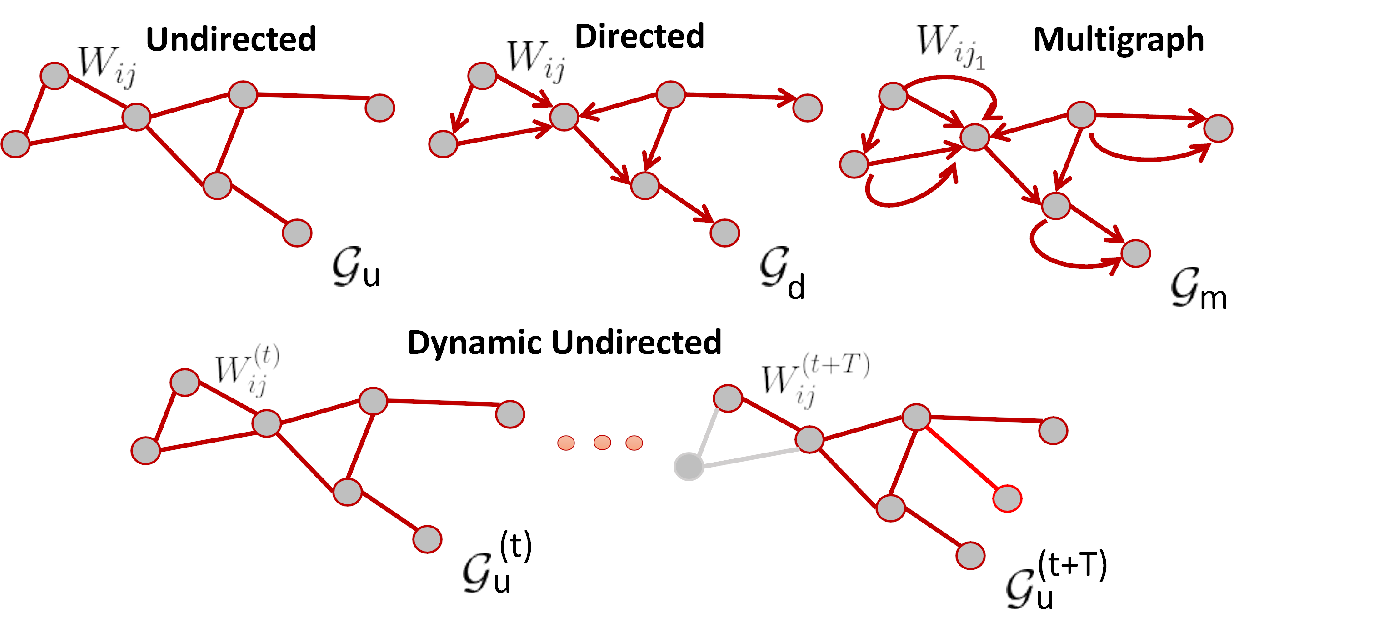}
    \caption{Types of graphs: above, static graphics; below, dynamic graphics;  $u$ denotes undirected, $d$ denotes directed, $m$ denotes multigraph, and $\mathcal{G}_u^{(t)}$ denotes a dynamic undirected graph at time $t$. $T{\in}\mathbb{N}$ represents a time offset.}
    \label{fig:types_graphs}
\end{figure*}

In addition, there are two variants of graphs that can be applied in sensor networks: \textit{i)} \textit{static} graphs $\mathcal{G}$ and \textit{ii)} \textit{dynamic} or time-varying graphs $\mathcal{G}^{(t)}$ whose structure depends on time $t$ \cite{skarding2021foundations}. Static graphs can be used to represent sensor networks that are static or fixed so the relationships between the sensors are not expected to change. In contrast, dynamics graphs have a time-dependent structure, meaning that the relationships between the sensors as well as the set of sensors composing the graph can change over time. This approach is useful to represent mobile sensor networks or dynamic networks where sensors can join and leave the network and the relationship between them can change. We can further classify the graphs into three types; \textit{i)} \textit{undirected}, in which the relationships between nodes are reciprocal, i.e., $(i,j){\in}\mathcal{E}$ implies $(j,i){\in}\mathcal{E}$, or $S_{ij}{=}S_{ji}$, \textit{ii)} \textit{directed}, in which the relationships do not need to be reciprocal, and \textit{iii)} \textit{multigraphs}, in which more than one edge can exist for a given pair of nodes $(i,j)$. Figure \ref{fig:types_graphs} represents the different types of graphs. In the specific case of monitoring sensor networks, where a graph structure is used to represent the relationship between the sensors' data, it makes sense to assume that the relationships between the sensor measurements are reciprocal, e.g., as in the case of correlation coefficients. This survey therefore focuses on the use of undirected static graphs, as monitoring platforms are typically fixed after deployment. More complex scenarios, such as mobile monitoring sensor networks, require more complex techniques and the use of dynamic graphs\footnote{Refer to \cite{yan2024signal} for a systematic review of dynamic graphs.}.

\begin{figure}[!h]
    \centering
    \includegraphics[width=0.55\columnwidth]{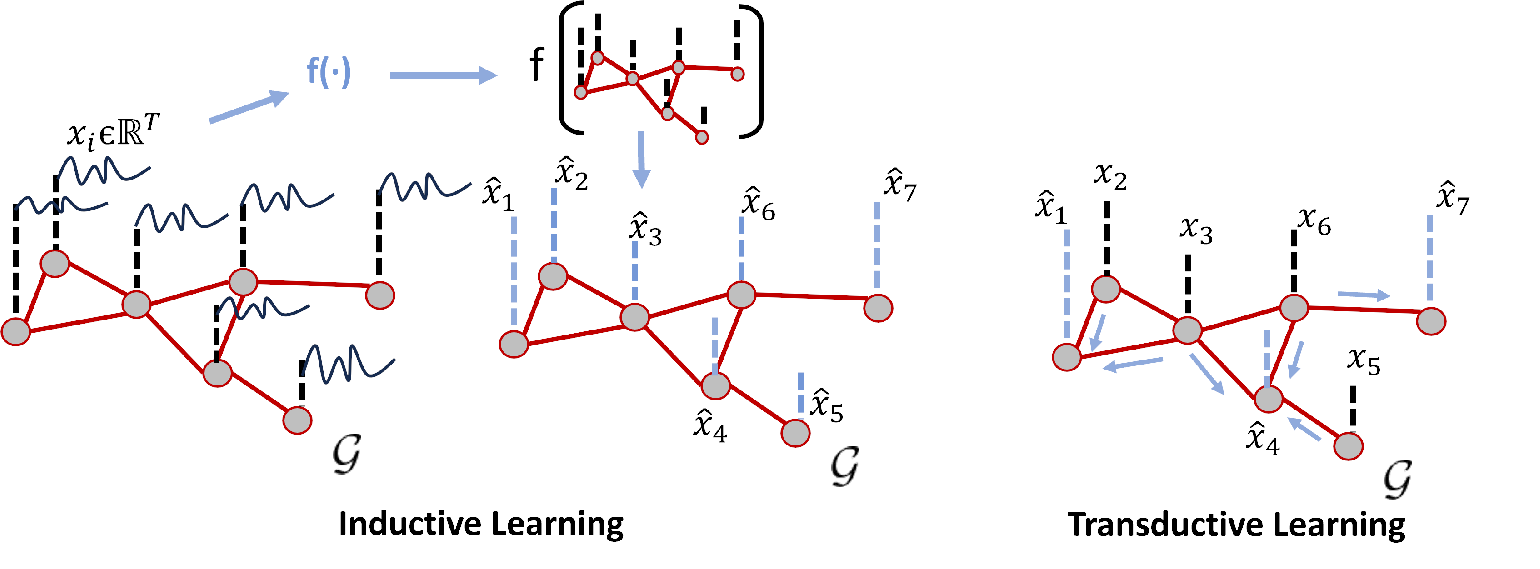}
    \caption{Inductive models are trained to perform inferences in other test sets while transductive models are trained to perform inferences for a specific test set.In black training data, and in blue testing data.}
    \label{fig:inductive}
\end{figure}

In addition, before moving on to explain the technical aspects of signal processing and ML over graphs, it is important to highlight the use of two types of predictive models in this graph-based approach; \textit{i)} \textit{inductive} models, and \textit{ii)} \textit{transductive} models. The objective of inductive learning is to infer a model from training instances, then this model can be applied to different testing instances. In contrast, in the case of transductive learning, training instances are used to obtain estimates for specific test instances. As we will see later, even though in the realm of sensor networks many models can be used in both ways, there are specific models of both types, such as a graph filter (inductive) or a graph kernel ridge regression (transductive). Figure \ref{fig:inductive} illustrates the idea behind these two learning paradigms.

\subsection{Graph Signal Processing}
\begin{figure}[!h]
    \centering
    \includegraphics[width=0.7\textwidth]{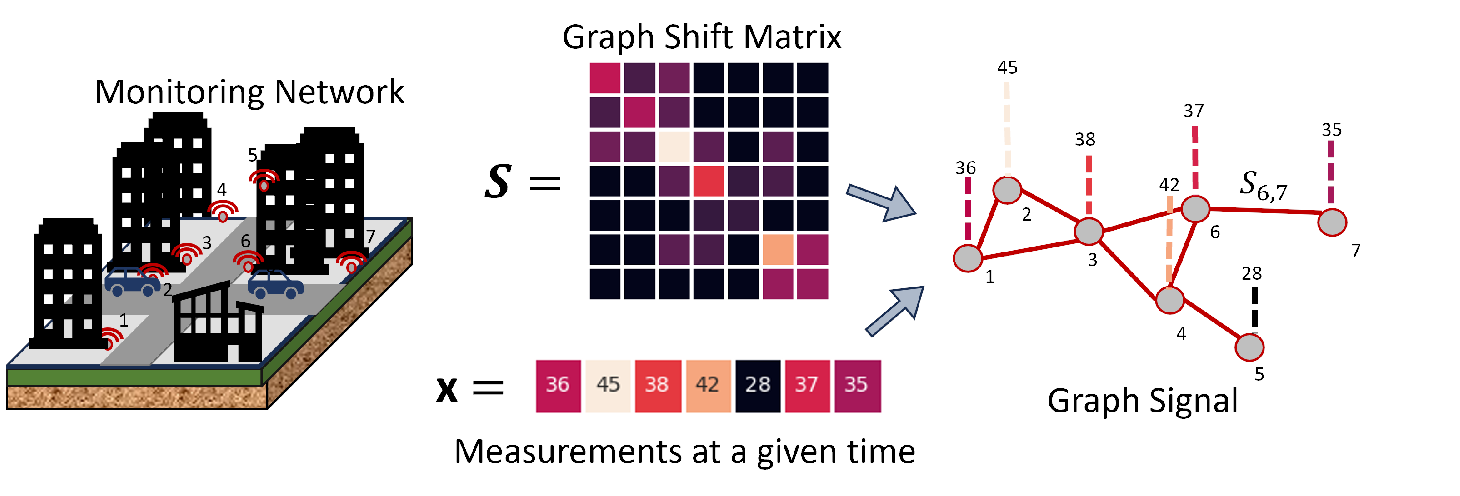}
    \caption{The GSP framework defines sensor network measurements as signals defined over graphs.}
    \label{fig:intro_GSP}
\end{figure}

The GSP emerged to adapt the signal processing foundations to signals with irregular support such as graphs \cite{sandryhaila2013discrete, shuman2013emerging, ortega2018graph}. A graph signal is defined as the map $x{:\;} \mathcal{V}\rightarrow\mathbb{R}$, then the $i$-th sensor measurement at a given time instant can be expressed as $x_i(t){\in}\mathbb{R}$ and the collection of measurements of the network at a given instant as $\mathbf{x}(t){\in}\mathbb{R}^N$ and a collection of measurements for a period of time as $\mathbf{X}{=}[\mathbf{x}(t_1), \dots, \mathbf{x}(t_T)]{\in}\mathbb{R}^{N\times T}$. One of the key components of this framework is the graph shift matrix $\mathbf{S}$ which describes the graph topology, and allows us to define the notion of graph shift, the GDFT, as well as the notion of signal smoothness among others. From now on, we simplify the notation of a graph signal at a given instant as $\mathbf{x}(t)\equiv\mathbf{x}$. Figure \ref{fig:intro_GSP} illustrates the use of the GSP framework to model the measurements of a sensor network, where the measurements are defined as signals on the graph and the graph shift matrix $\mathbf{S}$ describes the relationships between the sensors.

\subsubsection{Shift Matrix}
\label{subsubsec:shift}
\begin{figure}
    \centering
    \includegraphics[width=0.55\columnwidth]{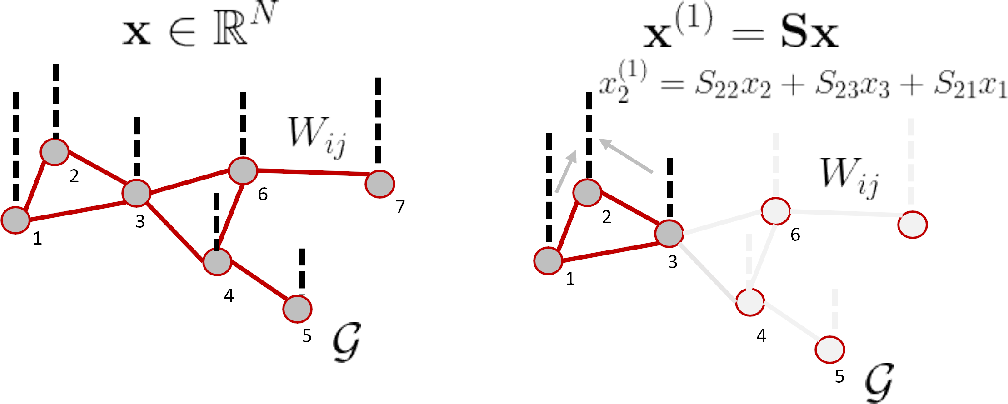}
    \caption{Representation of a graph signal $\mathbf{x}{\in}\mathbb{R}^N$ over a graph $\mathcal{G}$ and the effect of the graph shift operation via the graph shift matrix $\mathbf{S}$. The graph shift operator can be seen as a linear combination of the graph signal evaluated at neighboring nodes weighted by the entries of the shift matrix.}
    \label{fig:gsp}
\end{figure}

The entries of the graph shift matrix $S_{ij}$ are used to describe the relationship between the different pairs of graph nodes ($i$, $j$). The graph shift matrix is used as signal shift operator \cite{gavili2017shift}:
\begin{equation}
    \mathbf{x}^{(1)}=\mathbf{S}\mathbf{x}
\end{equation}
Hence, the graph shift is defined as the linear combination of the components of a graph signal $\mathbf{x}{\in}\mathbb{R}^N$ weighted by the entries of the shift matrix, Figure \ref{fig:gsp}. There exist different representations for the  graph shift matrix $\mathbf{S}$, among the most common we find; \textit{i)} graph \textit{adjacency} matrix $\mathbf{A}{\in}\mathbb{B}^{N\times N}$ which indicates the existing edges in the graph ($\mathbf{S}{=}\mathbf{A}$), such that $A_{ij}{=}1$
iff $(i,j){\in}\mathcal{E}$ and $A_{ij}{=}0$ otherwise, \textit{ii)} the graph \textit{weight} matrix $\mathbf{W}{\in}\mathbb{R}^{{+}^{N\times N}}$ which assigns a non-negative weight to the graph edges ($\mathbf{S}{=}\mathbf{W}$), i.e., $W_{ij}{>}0$ iff $(i,j){\in}\mathcal{E}$, and \textit{iii)} the combinatorial \textit{Laplacian} matrix $\mathbf{L}{\in}\mathbb{R}^{{N\times N}}$ ($\mathbf{S}{=}\mathbf{L}$), that is defined as $\mathbf{L}{=}\mathbf{D}{-}\mathbf{W}$, where $\mathbf{D}{=}diag(\mathbf{W}\mathbf{1})$, where $\mathbf{1}{\in}\mathbb{R}^N$ is a vector of ones. In fact, the graph shift matrix expresses the dependencies and independencies between graph nodes and it is usually forced to be sparse in order to capture the most critical relationships within the graph and to take advantage of the computational benefits of manipulating sparse matrices \cite{bunch2014sparse}. Other graph shift matrices can be found in \cite{shuman2013emerging}.

To be able to define graph filters, we need to adapt the notion of classical signal frequency spectrum to the graph domain. For this purpose, the GDFT is used to provide the frequency spectrum of a graph signal via the graph shift matrix eigendecomposition \cite{ricaud2019fourier}. The eigendecomposition of matrix $\mathbf{S}$ is defined as:
\begin{equation}
    \mathbf{S} = \mathbf{U}\boldsymbol{\Lambda}\mathbf{U}^{-1}
\end{equation}
where the eigenvector matrix $\mathbf{U}{=}[\mathbf{u}_1, \dots, \mathbf{u}_N]{\in}\mathbb{R}^{N\times N}$ defines the components of the Fourier modes and $\boldsymbol{\Lambda}{=}diag([\lambda_1, $ $\dots, \lambda_N]){\in}\mathbb{R}^{N\times N}$ includes the frequencies defined by the eigenvalues. For undirected graphs, the matrix $\mathbf{S}$ is real symmetric so $\mathbf{U}^{-1}{=}\mathbf{U}^{\mathsf{T}}$. Slow-varying components are associated with small eigenvalues while fast-changing components are associated with large eigenvalues. In this way, the GDFT can be defined as \cite{sandryhaila2013discretefourier}:
\begin{equation}
    \hat{\mathbf{x}}=\mathbf{U}^{-1}\mathbf{x}
\end{equation}
where $\hat{\mathbf{x}}{\in}\mathbb{R}^N$ is the GDFT transform of $\mathbf{x}$. This aspect will be further reviewed in the next section \ref{subsubsec:graph_filters}. As we have seen, the graph shift matrix is a core element in this graph-based approach, because of that, there exists extensive literature tackling the graph learning problem \cite{mateos2019connecting, dong2019learning}. The goal of the graph learning task is to construct a graph, i.e., to build the matrix $\mathbf{S}$ (which in turn defines the set of edges $\mathcal{E}$), that best describes the relations in the network. Here, we enumerate the two most common approaches:
\begin{itemize}
    \item \textit{Graph based on prior information}:  the graph is constructed using some prior information. An example can be the use of the geographical distance, where a weighting function based on the nodes' geographical distance is used \cite{sandryhaila2013discrete}. For instance, the Gaussian radial basis function is commonly used to assign weights, $W_{ij}{=} e^{\frac{-d_{ij}}{2\sigma}}$, where $d_{ij}{\in}\mathbb{R}$ is the geographical distance between nodes $i$ and $j$. Then, a threshold $TH{\in}\mathbb{R}$ can be used to force a certain graph sparsity. Another commonly used approach to restricting the sparsity of the resulting graph is building a K-nearest neighbor approach, i.e., only keeping the K most important edges per node. Some other examples may include the construction of the adjacency matrix according to the connectivity between sensors, i.e., $A_{ij}{=}1$ iff sensors $i$ and $j$ have communication.

    \item \textit{Data-driven graph}:  the graph shift $\mathbf{S}$ is learned from a given set of graph signals $\mathbf{X}{\in}\mathbb{R}^{N\times T}$ so that the resulting graph fits the collected measurements. The general data-driven graph learning problem can be defined as an optimization problem:
    \begin{align}
        \min_{\mathbf{S}\in\mathbb{R}^{N\times N}} & L(\mathbf{X}, \mathbf{S}) + \lambda \cdot R(\mathbf{S}) \\
         s.t. \;\; & \mathbf{S} \in \mathcal{S}
    \end{align}
    where the function $L(\mathbf{X}, \mathbf{S})$ evaluates the goodness-of-fit of the graph with respect to the set of measurements $\mathbf{X}$ and $R(\mathbf{S})$ is a regularization function that controls the complexity of the solution and is governed by a hyperparameter $\lambda{\in}\mathbb{R}$. The constraint forces the graph shift matrix $\mathbf{S}$ to belong to the set of valid graph shift matrices $\mathcal{S}$, whose properties depend on the specific matrix used, e.g., Laplacian matrix, normalized Laplacian matrix, weight matrix, etc.

    Different graph learning problems have been formulated, some of them make use of the signal smoothness criterion as function $L(\cdot)$, and others use different regularization functions $R(\cdot)$ to force a specific graph connectivity. For instance, Dong \textit{et al.} \cite{dong2019learning} review different graph learning methods, from latent factor models to models based on graph filters and graph dictionaries. In fact, Dong \textit{et al.} \cite{dong2016learning} develop a factor analysis-based model to learn the Laplacian matrix $\mathbf{L}$ using the smoothness criterion along with a Frobenius norm. Kalofolias \cite{kalofolias2016learn} propose an optimization model to learn the graph shift matrix $\mathbf{W}$. Other examples include the learning of valid precision matrices, or GNNs to learn a graph representation and a GNN model jointly \cite{egilmez2017graph, xia2021graph}. For instance, Ferrer-Cid \textit{et al.} \cite{ferrer2021graph} compare the use of data-driven (graph signal processing-based and statistics-based) and distance-based graph learning methods for air quality monitoring networks. Moreover, algorithms have been developed to learn time-varying graphs, allowing the graph to change over time according to the set of samples \cite{mateos2019connecting}.

\end{itemize}

This graph learning task plays an important role in IoT monitoring sensor networks, where the graph is usually fed to an optimization model to carry out a specific task, e.g., missing value imputation. Thus, the correct identification of the graph structure, i.e., the relationships within the data, can impact the performance of the final task.

\subsubsection{Signal Smoothness}
Signal smoothness is a widely adopted concept in optimization problems and applications. Broadly speaking, a graph signal $\mathbf{x}{\in}\mathbb{R}^N$ is said to be smooth with respect to an underlying graph if the graph strongly connects nodes with similar measurements and dissimilar nodes are weakly connected or disconnected. Formally, the signal smoothness is usually evaluated through the quadratic form \cite{stankovic2020datapart1}:
\begin{equation}
    S(\mathbf{S}, \mathbf{x})=\mathbf{x}^{\mathsf{T}}\mathbf{S}\mathbf{x}
\end{equation}
This metric has been widely used in problems such as graph learning, outlier detection, and missing value imputation \cite{gopalakrishnan2019identification, dong2016learning}. In the case $\mathbf{S}{=}\mathbf{L}$ then $S(\mathbf{L}, \mathbf{x})=\frac{1}{2}\sum_{i,j{\in}\mathcal{V}}W_{ij}(x_i - x_j)^2=$ meaning that the smaller the Laplacian quadratic form the more smooth the signal is. Nevertheless, we must place special care when using this criterion since its minimum is achieved for a total disconnected graph $\mathcal{E}{=}\varnothing$ or a constant or null graph signal $\mathbf{x}{=}\mathbf{0}$. Analogously to classic signal processing, the graph TV can be defined through the definition of the graph shift operator $\mathbf{S}$ \cite{sandryhaila2014discrete}:
\begin{equation}
    \text{TV}_p(\mathbf{S}, \mathbf{x}) = \|\mathbf{x} - \mathbf{S}\mathbf{x}\|_p
\end{equation}
where $p{\in}\mathbb{N}^+$ and $\|\cdot\|_p$ denotes the $l^p$ vector norm. This metric has also been used for signal reconstruction as well as for graph signal classification since it is a global graph metric, i.e., signal-wise \cite{chen2015signal}.

Other metrics exist to evaluate the variation of a graph signal, such as the local variation of a graph signal or the discrete p-Dirichlet energy form. Recently, the Sobolev smoothness for graph signals has been used for applications such as signal reconstruction for missing value imputation in sensor networks. The Sobolev smoothness for an undirected graph can be defined as \cite{giraldo2022reconstruction}:
\begin{equation}
    S_{\beta,\epsilon}(\mathbf{L}, \mathbf{x}) = \mathbf{x}^{\mathsf{T}}(\mathbf{L} + \epsilon\mathbf{I})^\beta\mathbf{x}
\end{equation}
where $\epsilon{\in}\mathbb{R}$ and $\beta{\in}\mathbb{R}$ are hyperparameters. In fact, the minimization of a graph signal smoothness can be translated to the penalization of the GDFT components $\hat{\mathbf{x}}_i$ according to the frequencies $\lambda_i$, being the largest frequencies the most penalized ones \cite{giraldo2022reconstruction}.

\subsubsection{Graph Filters}
\label{subsubsec:graph_filters}
Now that we have explained the use of the graph shift matrix as well as the utility of its eigendecomposition to formulate the GDFT, we can formulate the graph filters. We assume that the eigenvectors $\mathbf{u}_i$ are ordered in ascending order according to the eigenvalues, $\lambda_1{\leq} \dots {\leq} \lambda_N$. Given the definition of the GDFT \cite{sandryhaila2013discretefourier}, a non-parametric filter can be defined as a function of the eigenvalues in frequency domain, or a function of the shift matrix in vertex domain \cite{sandryhaila2014discrete}:
\begin{align}
\tilde{\hat{\mathbf{x}}} = & g(\boldsymbol{\Lambda})\hat{\mathbf{x}} \\
\tilde{\mathbf{x}} = &  \mathbf{U}g(\boldsymbol{\Lambda})\mathbf{U}^{-1}{\mathbf{x}} = f(\mathbf{S})\mathbf{x}
\end{align}
More precisely, a linear graph filter can be defined in the vertex domain as \cite{sandryhaila2013discrete}:
\begin{equation}
    \tilde{\mathbf{x}} = \sum_{i=0}^{K-1} h_i\mathbf{S}^i \mathbf{x}
\end{equation}
where $\mathbf{h}{\in}\mathbb{R}^K$ are the filter's taps. The filtered version $\tilde{\mathbf{x}}$ of the signal $\mathbf{x}$ is a linear combination of the shifted versions of the signal, $\mathbf{S}^k\mathbf{x}$. Other filters can be used in the vertex domain to carry out applications such as denoising, signal reconstruction, or anomaly detection, as well as frequency-vertex domain representations using graph wavelet transforms or graph slepians \cite{xiao2020nonlinear, van2017slepian}. In fact, there exists a wide variety of filters, from linear to nonlinear, which allow the implementation of powerful predictive models capable of reconstructing the signal at any of the sensors of a network. Examples of filters include convolutional filters \cite{wu2019simplifying}, nonlinear polynomial graph filters \cite{xiao2020nonlinear}, Volterra-like filters \cite{xiao2021distributed}, etc.

To approximate the computation of the polynomials of $\mathbf{L}$ and $\boldsymbol{\Lambda}$ for a specific transfer function, different bases can be used, e.g., the Chebyshev polynomials, the Gegenbauer polynomials, and the monomial or Bernstein basis among others \cite{he2021bernnet, puny2023equivariant, castro2024gegenbauer}. These approximations provide a recursive solution to approximating the desirable transfer functions providing computational benefits. Most graph filters allow for a distributed implementation given the use of graph shifts, which can be seen as a result of a diffusion process where only connected nodes may share information \cite{coutino2019advances}. This feature is particularly interesting in the field of sensor networks where edge and fog computing approaches can be employed to implement a specific data-driven model.

The notion of bandlimited signals has been widely used to develop signal reconstruction models as well as sensor sampling schemes \cite{huang2020reconstruction}. We can define the set of $K$-bandlimited graph signals as $\mathcal{K}{=}\{\mathbf{x}{:\;} \hat{\mathbf{x}}_i{=} 0 \;\text{for}\; K{\leq} i{\leq} N\}$. Many reconstruction algorithms rely on the assumption that a given graph signal has a sparse representation in the spectral domain. Besides, sensor sampling selection schemes have been developed to find the subset of sensor measurements that produce the lowest signal reconstruction error \cite{rusu2017node}.

To conclude this section, we delve into the graph convolution operation, which plays an important role in graph convolutional neural networks (section \ref{subsec:gnn}). It is well-known that the graph signal convolution $\mathbf{x}\ast\mathbf{f}$ is equivalent to the multiplication in the spectral domain \cite{stankovic2020datapart3}:
\begin{equation}
\label{eq:conv}
    \mathbf{x}\ast\mathbf{f} = \mathbf{U}\left ((\mathbf{U}^{-1}\mathbf{x})\odot(\mathbf{U}^{-1}\mathbf{f}) \right )
\end{equation}
where $\mathbf{x}{\in}\mathbb{R}^N$ and $\mathbf{f}{\in}\mathbb{R}^N$ are both graph signals and $\odot$ corresponds to the Hadamard product. Thus, the convolution is equivalent to the multiplication of the spectral components by a convolution signal $\mathbf{f}$. Recalling the previous equation \ref{eq:conv} we can observe how a graph filter in the spectral domain can be seen as a convolution operation:
\begin{equation}
    \mathbf{x}\ast g= \mathbf{U}g(\boldsymbol{\Lambda})\mathbf{U}^{-1}\mathbf{x} = \mathbf{U}\begin{bmatrix}
g(\lambda_1) & \dots& 0 \\
 \vdots& \ddots & \vdots\\
0& \dots& g(\lambda_N)
\end{bmatrix} \mathbf{U}^{-1}\mathbf{x}
\end{equation}
Thus, this resembles a graph filter where the polynomials of $\mathbf{L}$ and $\boldsymbol{\Lambda}$ are used to provide good localization properties in the vertex domain. Moreover, approximations such as the Chebyshev polynomials can be used to provide computational benefits \cite{defferrard2016convolutional}. This aspect will be further described in section \ref{subsec:gnn}, where spectral GNNs are introduced.

\subsection{Machine Learning over Graphs}
\label{subsec:mlog}
The use of ML on data residing in a manifold represented by a graph has been in use for years \cite{song2022graph, subramanya2022graph, belkin2004semi}. In the field of graph-based semi-supervised learning, data is assumed to reside on a manifold, with data points being related to each other, so that models can input such structured information to assist ML models. Other examples of ML tasks include the inference of node embeddings from graph data \cite{abu2018watch}. In this context, instead of working directly with graph-structured data, node embeddings are obtained and classic ML models are fed with these node representations.

In this section, we delve into the main uses of ML on graphs in sensor networks, which are the development of graph-regularized models and the interpretability they provide. Although many tasks and techniques could be included in graph ML, e.g., obtaining metrics, graph representation learning, node embeddings, or community detection, in this paper we focus on its use mainly through regularization and kernels in predictive models given its potential applicability to data quality in sensor networks.

\subsubsection{Graph-Regularized ML}
\begin{figure*}
    \centering
    \includegraphics[width=0.5\columnwidth]{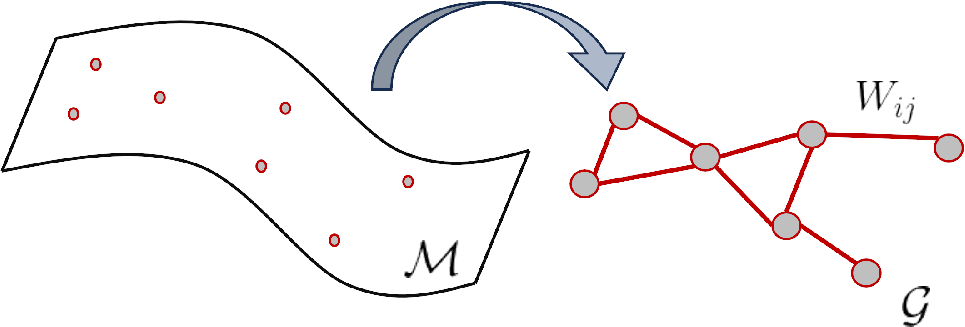}
    \caption{Representation of data points that lie in a manifold $\mathcal{M}$ and their representation in a graph form $\mathcal{G}$.}
    \label{fig:manifold}
\end{figure*}

In many cases, it is assumed that the data resides in a non-Euclidean space, such as a smooth manifold or a Riemannian manifold \cite{belkin2004semi, zhou2014semi}. Hence, a graph can be used to describe the similarities of the data in such a manifold. Graph ML models aim to take advantage of a graph that describes the relationship between the data. This concept has been widely used in graph semi-supervised learning, regularization, and kernels over graphs. Thus, a manifold $\mathcal{M}$ is approximated via a graph $\mathcal{G}$, Figure \ref{fig:manifold}.

The Laplace-Beltrami operator of functionals on Riemannian manifolds is widely used and the graph Laplacian operator can be seen as its discrete approximation when the number of data points tends to infinity \cite{hu2021graph, burago2015graph, sandryhaila2013discrete}. Then, the smoothness of a function over a graph is represented by the quadratic form $\mathbf{x}^{\mathsf{T}}\mathbf{L}\mathbf{x}$, which is akin to the graph signal smoothness \cite{dong2019learning, belkin2004semi}.
Now, we can illustrate the use of graph-based regularization in a regression task, independently of whether it is supervised or semi-supervised, as finding the map $f{:\;}\mathbb{R}^M\rightarrow\mathbb{R}^N$ where $N{>}M{\geq} 1$:
\begin{equation}
    \min_f \text{Loss}\left (\mathbf{y}, \mathbf{m}\odot f(\mathbf{x})\right ) + \gamma \cdot R_{\mathcal{G}}(f)
\end{equation}
where $\mathbf{x}{\in}\mathbb{R}^M$, $\mathbf{y}{\in}\mathbb{R}^M$, $\mathbf{m}{\in}\mathbb{B}^N$ is a selector vector that selects the observed entries $\mathbf{x}$ of the graph signal $f(\mathbf{x})$, $\text{Loss}(\cdot)$ is a loss function, $\gamma{\in}\mathbb{R}$ is an hyperparameter promoting the regularization, and $R_{\mathcal{G}}(f)$ is a graph-based regularizing function. An example of regularization is the well-known Tikhonov regularization for graphs where the estimated graph signal $f(\mathbf{x})$ is forced to be smooth with respect to the graph $\mathcal{G}$, the loss function is set to the residual sum of squares and the regularization term is set to the Laplacian quadratic form \cite{yang2021node}:
\begin{equation}
 \min_f \underbrace{\| \mathbf{x} - \mathbf{m}\odot f(\mathbf{x}) \|_2^2}_{\text{Loss}\left (\mathbf{x}, f(\mathbf{x})\right ) }  + \gamma  \cdot\underbrace{{f(\mathbf{x})}^{\mathsf{T}}\mathbf{L}f(\mathbf{x})}_{R_{\mathcal{G}}(f)}
\end{equation}
This is an example of a transductive method, where for a given graph signal with some observed values, the signal is reconstructed taking into account the resulting signal smoothness. This formulation has been widely used for the regression of graph signals in scenarios such as image processing and sensor networks \cite{belkin2004semi, venkitaraman2019predicting, pilavci2021graph}. Similarly, in the context of supervised ML and semi-supervised ML, the norm in the Reproducing Kernel Hilbert Space (RKHS) $\mathcal{H}$ associated to a graph $\mathcal{G}$ has been used in Laplacian regularized least squares and in Laplacian support vector machines \cite{melacci2011laplacian}, $R_{\mathcal{G}}(f) {\equiv} \|f\|^{\mathcal{G}}_{\mathcal{H}}$.

More remarkably, the graph counterpart of RKHS has been used in graph ML \cite{romero2016kernel, jian2023kernel, venkitaraman2019predicting, sahbi2021kernel}. Graph kernels play an important role in ML for comparing graphs in tasks such as graph classification and sub-graph matching in fields such as biology \cite{borgwardt2020graph,vishwanathan2010graph}. In sensor networks, however, we are more interested in graph kernels that compare the relationships between nodes in a graph, $k{:\;}\mathcal{V}\times\mathcal{V}\rightarrow\mathbb{R}$. In this sense, we can define a graph kernel matrix $\mathbf{K}_{\mathcal{G}}$ that depicts the similarities between graph nodes as \cite{romero2016kernel}:
\begin{align}
    \mathbf{K}_{\mathcal{G}}  & = r^{\dagger}(\mathbf{L}) = \mathbf{U}r^{\dagger}(\boldsymbol{\Lambda})\mathbf{U}^{-1}
\end{align}
where $r{:\;}\mathbb{R}\rightarrow\mathbb{R}_+$ is a non-negative map and $\dagger$ denotes the pseudoinverse. Similarity functions such as graph diffusion or random-walk kernels have been used for graphs. This concept led to the development of graph kernel-based regression where the RKHS, $\mathcal{H}$, is defined for graph signals and its norm is used for regularization \cite{romero2016kernel}:
\begin{equation}
    R_{\mathcal{G}}(f) = \|f\|_{\mathcal{H}}^2 = <f, f>_{\mathcal{H}} = \boldsymbol{\alpha}^{\mathsf{T}}\mathbf{K}_{\mathcal{G}}\boldsymbol{\alpha}
\end{equation}
where $\boldsymbol{\alpha}{\in}\mathbb{R}^N$ is the set of learnable parameters. Then, from the representer theorem we can derive the estimate for the value of the signal at the different nodes \cite{romero2016kernel}. This concept has been used for kernel-based graph regression as well as Gaussian processes over graphs, kernel-kriged Kalman filters over graphs, and the inference of functions over graphs, using both graph kernels and kernels agnostic to the graph \cite{venkitaraman2020gaussian, ioannidis2018inference, romero2016kernel, smola2003kernels}.

Finally, many other applications have benefited from graph kernels, e.g., the factorization of matrices by introducing prior information as vertex-based graph kernels \cite{pal2018kernelized}. Other examples include the introduction of graph kernels for nonnegative matrix factorization \cite{li2023adaptive}, or the introduction of graph-regularizers in concept factorization techniques \cite{mu2023dual}.

\subsubsection{Interpretable ML}
The interpretability of ML and artificial intelligence models has been the focus of research in recent years given the increasing development and adoption of highly complex models such as neural networks (commonly known as black-box models). The interpretability of a ML model aims to reason why a prediction is made and to give further insights into the performance and functioning of a model \cite{molnar2020interpretable, kok2023explainable}.

Interpretable ML promotes techniques that are interpretable, such as generalized linear regression, generalized additive models, and decision trees among others \cite{du2019techniques}. Among the most common approaches to interpretability are; \textit{i)} model-specific interpretability, i.e., interpreting the inferred parameters, such as the coefficients of a linear regression, \textit{ii)} sparsity-promoting models to force feature selection and enhance the interpretability of the parameters, and \textit{iii)} model-agnostic interpretability, i.e., providing black-box models with tools to explain the results, such as neural network unrolling, feature visualization, latent embedding inspection, or results inspection.

Signal processing and ML over graphs have been shown to enhance the interpretability of models \cite{dong2020graph}. First, a graph provides an explicit structure that indicates the relationships between the sensors as well as the importance of the relationships. Thus, the graph shift matrix defines a node's neighborhood $\mathcal{N}(i){=}\{j{:\;} j{\in}\mathcal{V} \wedge S_{ij}{\neq} 0\}$, i.e., the set of influential nodes for a given node. In this line, in graph attention networks (GAT) the attention mechanism defines the importance of the nodes' features (or hidden features) in a similar manner as the graph shift matrix but allows attention at different levels, e.g., different layers and multi-head attention mechanism. The graph learning field seeks to identify the network structure, i.e., relationships between nodes, by promoting smoothness and frequency-domain constraints, therefore, improving the interpretability of the resulting graph. Furthermore, GSP has been used to enhance the interpretability of classic neural networks and graph neural networks by imposing graph-based regularization in hidden layers, forcing the hidden features to adapt to a given graph \cite{tong2020interpretable}. In this line, graph filters and graph signal denoising techniques provide an interpretation in the frequency domain, enhancing the interpretability of the results \cite{dong2020graph}. In fact, GNNs have been interpreted as the result of signal denoising and filtering processes \cite{chen2021graph}. Finally, it is worth noting that kernels over graphs and the use of kernelized graph methods also provide an interpretable alternative since the kernel map can denote similarities between nodes and similarities between graphs. The latter concept has been employed in the development of interpretable GNN using graph kernels \cite{feng2022kergnns}.

To sum up, a graph-based approach for sensor networks can provide insights into the relationships between sensors, just as the graph can be used to introduce prior information. Furthermore, depending on the application of the sensor network, if decision-making processes are to be fed with the network data it is important to be able to interpret the results. For instance, in the case of a network that monitors possible chemical leaks, it is important to be able to interpret the reasoning behind a detection as well as the leakage pattern using the graph.

\subsection{Graph Neural Networks}
\label{subsec:gnn}
The major limitation of well-known neural networks such as convolutional neural networks (CNNs) is their intrinsic application on data defined in regular grids such as images, tabular, or structured data. GNNs appeared as the translation of neural networks to data with irregular support such as graphs \cite{kipf2016semi,defferrard2016convolutional}. In this sense, most of the applications tackled by classic neural networks can be applied to GNNs by making use of the graph topology and the operators defined over it. For instance, spectral convolutional GNNs made use of the convolution operator to implement CNNs over data residing on graphs. Another related field is geometric deep learning, where broader geometric properties are taken into account, and tailored layers and techniques are applied to it, but, we do not cover this perspective in this paper\footnote{For more information about geometric deep learning refer to \cite{bronstein2017geometric}.}.

More precisely, we can broadly classify GNNs into \cite{wu2020comprehensive}; \textit{i)} graph convolutional neural networks (GCNs), \textit{ii)} graph autoencoders (GAEs), and \textit{iii)} graph recurrent neural networks. Spatiotemporal GNNs are also frequently included in this classification since they incorporate different building modules and may have tailored architectures. In this review, we will mention their main components without going into detail. More precisely, first, we will describe the building blocks of GNNs; \textit{i)} propagation, \textit{ii)} sampling, \textit{iii)} pooling and readout functions. Then, we will describe the main GNN architectures listed above, placing special emphasis on GCNs and their components given that these are widely used and can be used to implement other GNN architectures such as autoencoders. In the next sections, we explain the core elements of the different GNNs, and in section \ref{sec:applications} we review well-known architectures used for specific sensor network applications.

\subsubsection{GNNs: Building Modules}
\label{subsubsec:gnns_modules}
In this section, we explain the main components that allow the implementation of a broad range of architectures for different GNNs. Among the different components we find those defined in \cite{zhou2020graph}:

\begin{itemize}
    \item \textbf{Propagation Modules}: given that the data resides in a graph topology, propagation functions that propagate and process the values between different graph nodes are required. For instance, as we will explain in the next section, convolutional GNNs make use of convolution operators using spectral and spatial strategies to share and aggregate information between different graph nodes. Other propagation operators include recurrent operators such as gated recurrent units.

    \item \textbf{Sampling Modules}: are usually included in the propagation modules so that only relevant nodes or nodes' neighborhoods participate in the processing. This is especially interesting for large and highly dense graphs. Examples include the sub-sampling of nodes for each neighborhood, the sampling of edges, or even the sampling of sub-graphs via graph spectral clustering \cite{chiang2019cluster}.

    \item{\textbf{Pooling Modules}}: the goal of a graph pooling layer is to obtain coarser representations of the features or graphs. Sub-graphs can be identified and nodes and edges can be aggregated. This idea is related to the down-sampling notion of a signal where some coarser nodes are generated to represent a higher dimensional graph signal \cite{liu2022graph}. The most simple techniques for pooling are the simple $sum/mean/max$ functions \cite{henaff2015deep}:
    \begin{equation}
        \mathbf{h}^{(k)} = pool_{\mathcal{G}}\left (\textbf{h}^{(k)}_1, \cdots, \textbf{h}^{(k)}_{C_k} \right )
    \end{equation}
    where $\mathbf{h}_i^{(k)}$ represents the hidden features at the $k$-th layer and the $i$-th channel. Some other more sophisticated techniques, such as hierarchical pooling strategies, rearrange the graph nodes and create a new reduced set of nodes and edges representing the original ones \cite{bianchi2020spectral}. Some examples include graph coarsening and eigenpooling \cite{ma2019multi}. These techniques play an important role in scenarios where the output of the GNN is at graph level or edge level, meaning that some reduced graph representations might be required. Readout functions can be used to summarize final or intermediate representations of a graph so that new features can be obtained, e.g., aggregating hidden representations. These are particularly important in GNNs where the output is expressed at graph level, e.g., graph classification, so that the readout function produces a metric summarizing the entire graph \cite{wu2020comprehensive}. Moreover, readout functions can be used to produce node embeddings and aggregations so that classical, e.g. fully-connected or convolutional layers, can be used within a GNN.

\end{itemize}

\subsubsection{Convolutional GNNs}
\label{subsubsec:gcn}
Convolutional GNNs are commonly classified into \cite{wu2020comprehensive, zhou2020graph}; \textit{a)} spectral GNNs and \textit{b)} spatial GNNs. Spectral convolutional GNNs make use of the definition of graph signal convolution operator (section \ref{subsubsec:graph_filters}) to implement the classic convolution operation, while spatial convolutional GNNs make use of the graph structure to define nodes' neighborhoods and implement the convolution.

\begin{figure}[t]
    \centering
    \includegraphics[width=0.9\textwidth]{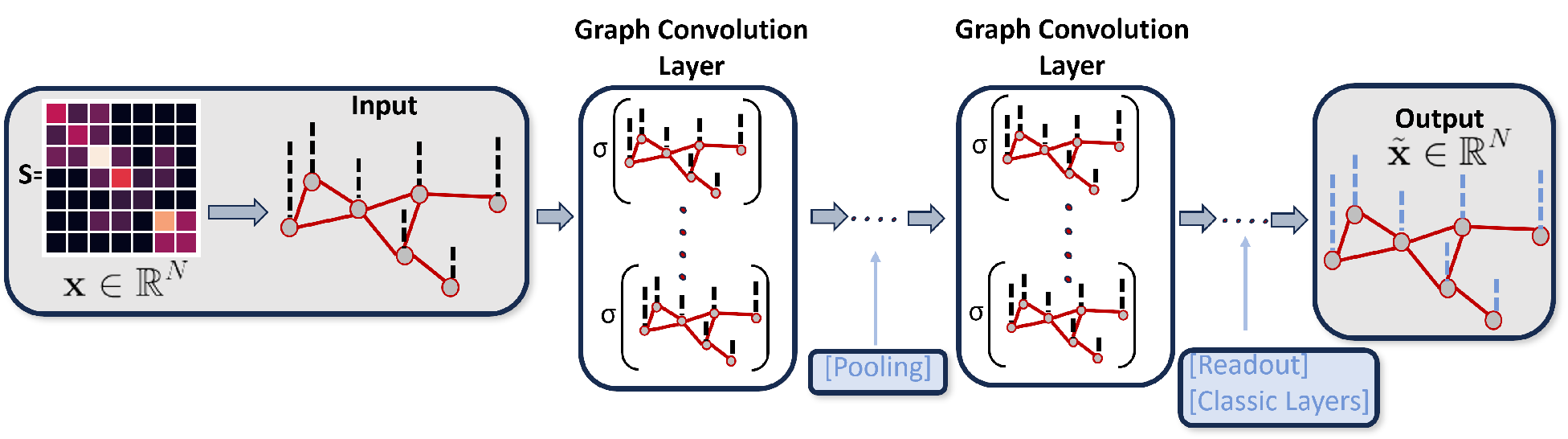}
    \caption{Example of a GCN architecture at the node level, where a graph signal $\mathbf{x}$ is reconstructed $\tilde{\mathbf{x}}$. The inputs are the graph signal and the graph shift matrix $\mathbf{S}$, and $\sigma(\cdot)$ denotes an activation function. A GNN architecture can include graph convolution and pooling layers, as well as readout layers and classic layers, e.g., fully-connected or convolutional layers.}
    \label{fig:gcn_example}
\end{figure}

\textit{\textbf{Spectral convolutional GNNs}}: make use of the graph convolution operator $\mathbf{x}\ast\mathbf{f}$ to define the convolution layers \cite{kipf2016semi, defferrard2016convolutional}. In a general manner, we can define a spectral graph convolution layer as:
\begin{equation}
\begin{split}
\mathbf{h}^{(k)}_{j} = \sigma \left (agg_{\ i=1,\dots, c_{k-1}}\left (\mathbf{h}_i^{(k-1)}\ast\boldsymbol{\theta}_{ij}^{(k)} \right ) \right ), \hspace{2em} j=1,\dots,c_k
\end{split}
\end{equation}
where $\mathbf{h}^{(k)}_{c_k}{\in}\mathbb{R}^N$ is the $c_l$ (channel) hidden state at layer $k$, $\sigma(\cdot)$ is the activation function, $c_k{\in}\mathbb{N}^+$ is the number of channels in layer $k$, $agg(\cdot)$ is an aggregation function used to aggregate the different filtered versions of graph signal $\mathbf{h}_i^{(k-1)}{\in}\mathbb{R}^N$ (e.g., sum or average), and $\boldsymbol{\theta}_{ij}^{(k)}{\in}\mathbb{R}^{K_{ij}}$ is the convolution function associated with a set of learnable parameters. Different types of networks and convolutions assume a different parametrization of the learnable parameters $\boldsymbol{\theta}$. For instance, the GCN \cite{kipf2016semi} and Chebnet \cite{defferrard2016convolutional} use the Chebyshev polynomials to approximate the filtering, the CayleyNet \cite{levie2018cayleynets} uses the Cayley polynomials to approximate the filtering, the Graph Wavelet Neural Network (GWNN) \cite{xu2019graph} uses wavelets to approximate the convolution operation, and the graph diffusion neural network (GDNN) \cite{gasteiger2019diffusion} uses a graph diffusion convolution to implement the message passing operation, combining strengths of spatial and spectral methods. For instance, the implementation of the spectral graph convolution using the truncated expansion of the Chebyshev polynomials results in \cite{defferrard2016convolutional}:
\begin{equation}
    \mathbf{x}\ast\boldsymbol{\theta} = \sum_{k=0}^{K-1}\theta_k T_k(\mathbf{L})\mathbf{x}
\end{equation}
where $\boldsymbol{\theta}{\in}\mathbb{R}^K$ are the polynomials coefficients and $T_k(\mathbf{L})$ corresponds to the recursive implementation of the truncated Chebyshev polynomials of $\mathbf{L}$ \cite{defferrard2016convolutional}. Figure \ref{fig:gcn_example} depicts a classic architecture for a convolutional GNN where different pooling modules and layers can be included.

\textit{\textbf{Spatial convolutional GNNs}}: in this case, the spatial notion in a graph $\mathcal{G}$ is used, where the graph structure explicitly tells which nodes are related (node neighborhood) and to what extent \cite{zhang2019graph, danel2020spatial}. Among the most representative spatial graph convolutional neural networks we find propagation-based models such as the diffusion-based graph convolutional neural network (DCNN) \cite{atwood2016diffusion}, the GraphSAGE \cite{hamilton2017inductive} or generalistic frameworks such as message-passing neural networks (MPNNs) \cite{gilmer2017neural}. Nonetheless, we can define a general spatial graph convolutional layer as:
\begin{align}
    &\mathbf{a}^{(k-1)}(n) = agg_{\forall i} \left ([\mathbf{h}^{(k-1)}_{i}(n) || \mathbf{h}^{(k-1)}_{i}(\mathcal{N}(n))], \boldsymbol{\theta}_{i}^{agg, (k)} \right ),  \nonumber  \\
    & \mathbf{h}^{(k)}_{j}(n) = f \left (\mathbf{h}_j^{(k-1)}(n), \mathbf{a}^{(k-1)}(n) , \boldsymbol{\theta}_{j}^{f, (k)} \right ) \hspace{1em}    ,j=1,\dots,c_k
\end{align}
where $f(\cdot)$ is a function that combines the value at node $n$ with a function of the vicinity, $agg_{\forall i}(\cdot)$ is a function that aggregates the information of neighboring nodes of $\mathcal{N}(n)$, and $||$ denotes a concatenation operation. Function $f$ combines the aggregated spatial information at node $n$ ($\mathbf{a}^{(k-1)}(n)$) with the previous hidden state $\mathbf{h}^{k-1}(n)$ at node $n$. Both $agg$ and $f$ can contain a set of learnable parameters $\boldsymbol{\theta}_{i}^{agg}$ and $\boldsymbol{\theta}_{j}^f$. This framework provides great generalization capabilities allowing the definition of aggregation and combination functions according to the data. The GAT \cite{velickovic2017graph} make use of the node spatial notion and graph structure to define the graph-based analogs of attention mechanisms.

\subsubsection{Autoencoder GNNs}
GAEs are widely used in applications such as outlier detection and missing value imputation. GAEs are trained to reconstruct the input signal, and in the process, a latent representation of the data is learned. To force the neural network to learn a latent representation, the dimensions of the hidden features of the encoder are decreased to produce a bottleneck effect, in what is called an undercomplete autoencoder, or the dimensions are increased, in what is called an overcomplete autoencoder. Hence, this idea is related to representation learning, where latent representations of the data are learned, either by obtaining a compressed version of the data or an overexpressed version. Variational graph autoencoders (VGAE) \cite{kipf2016variational} appeared to perform edge-level prediction tasks by making use of graph convolution layers to produce the latent representation of the data. In fact, graph convolution layers are the building blocks of GAEs, however, other layers can be used \cite{salha2019keep}. Graph variational autoencoders (GVAE) \cite{simonovsky2018graphvae} also make use of a GCN for the encoder. Graph autoencoders typically use a graph convolution approach to produce the latent features and the decoder part can use some well-known neural network layers to produce the output. An example is the marginalized graph autoencoder \cite{wang2017mgae} that combines the use of graph convolutional layers with spectral clustering. Adversarially regularized graph autoencoders (ARGA) \cite{pan2018adversarially} make use of a graph convolution-based encoder and a graph embedding-based decoder to perform link prediction. Dirichlet graph variational autoencoders (DGVAE) \cite{li2020dirichlet} use Dirichlet priors and spectral clustering to enhance variational autoencoders. All in all, GAEs can be implemented by means of the layers seen so far, combining ideas from representation learning and dimensionality reduction to perform tasks at different levels, e.g., signal reconstruction, and link prediction.

\subsubsection{Recurrent GNNs}
Recurrent GNNs usually combine GCNs with recurrent modules to take into account the temporal patterns of the data \cite{liu2022introduction}. For instance, graph convolutional recurrent neural networks (GCRNs) put LSTM and the ChebNet together. Diffusion convolutional recurrent neural networks (DCRNNs) \cite{li2017diffusion} combine diffusion graph convolutional layers with gated recurrent units. Gated graph neural networks (GGNNs) \cite{li2015gated} use gated recurrent units with spatial convolutions to implement the recurrence \cite{wu2020comprehensive}:
\begin{equation}
    \mathbf{h}^{(k)}(n) = \textbf{GRU}\left (\mathbf{h}^{(k-1)}(n), agg_{u\in\mathcal{N}(n)}\left (\mathbf{h}^{(k-1)}(u), \boldsymbol{\theta}\right )\right )
\end{equation}
where the hidden representation of node $n$ is a gated recurrent unit applied to the previous state of the $n$-th node ($\mathbf{h}^{k-1}(n)$) and a combination of the neighbors' previous representations $agg_{u\in\mathcal{N}(n)}(\mathbf{h}^{(k-1)}(u), \boldsymbol{\theta})$. $\boldsymbol{\theta}$ are learnable parameters. These are just some examples of recurrent neural networks and how these are implemented by means of convolution operations together with classic recurrent units.

Spatiotemporal graph neural networks combine both convolutional and recurrent layers to learn spatial and temporal dependencies, which can be encoded by time-varying graphs and signals. For instance, graph convolutions can be used to model spatial dependencies, and gated units or classic convolutional layers can be used to model temporal dependencies. Common applications include forecasting techniques \cite{li2021spatial, bui2022spatial}.

\subsubsection{Generative GNNs}
\label{subsubsec:gans}
Generative models have attracted attention in recent years. Specifically, generative models can generate new synthetic measurements, or more complex structures such as graphs and sub-graphs. There are two main classes of models that are the most widely used, the variational autoencoders and the generative adversarial networks (GANs). In the context of graph variational autoencoders, models have been developed that make use of GCNs as encoders and learn a latent generative model $q(h)$ (where $h$ is the latent representation of the data) that allows for generating new instances of the graph shift matrix or graph signals \cite{kipf2016variational,ahn2021variational}. For example, Do \textit{et al.} \cite{do2020graph} use a VGAE consisting of an encoder formed by stacked layers of GCNs, a latent model parametrized as a Gaussian distribution, and a decoder composed of stacked GCNs to complete air quality measurements. Another successful example of generative models are GANs in which a sample generator $g$ competes with a discriminator whose goal is to distinguish real instances from generated instances in a zero-sum game \cite{pan2018adversarially,wang2018graphgan}. For example, VGAEs can be used as generators to generate synthetic samples or graphs \cite{pan2018adversarially} or more customized techniques can be developed to generate possible graph shift matrices \cite{wang2018graphgan}.

Throughout this section, we have reviewed the foundations of GSP, ML over graphs and GNNs. In the next section, we review articles that apply many of the fundamentals we have seen to improve data quality in the context of monitoring sensor networks using graph models.

\section{Graph-Based Data Quality Applications}
\label{sec:applications}

\begin{figure*}[!t]
    \centering
    \includegraphics[width=0.45\columnwidth]{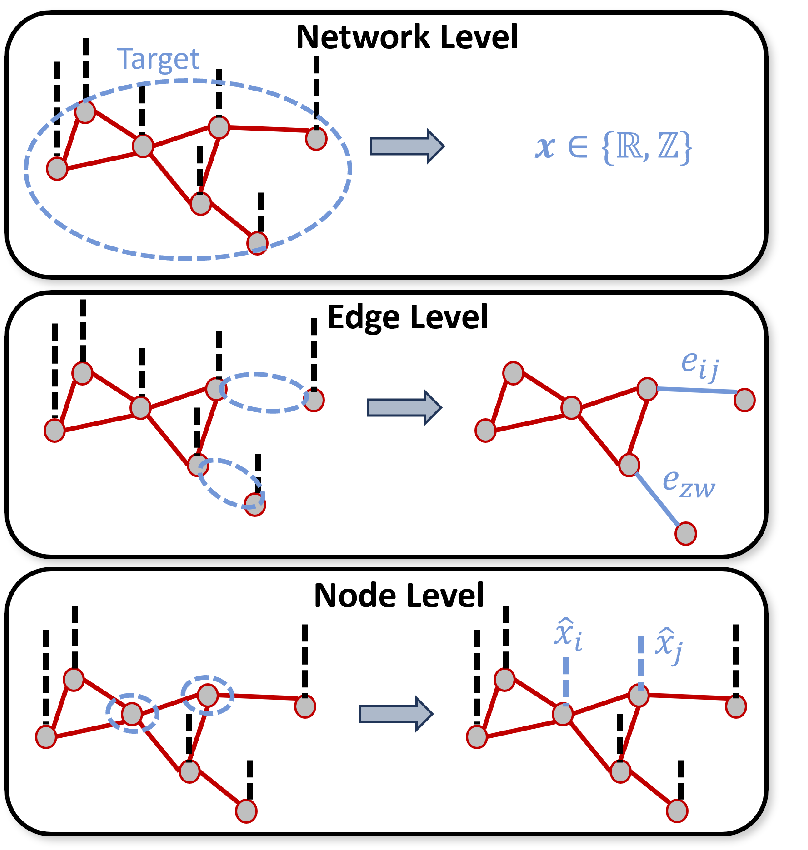}
    \caption{Different task levels for graph-based techniques in monitoring sensor networks; network level tasks, edge level tasks, and node level tasks.}
    \label{fig:tasks_levels}
\end{figure*}

As we have reviewed in previous sections, many graph-based techniques allow the modeling of sensor network measurements \cite{dong2023graph}. There is a wide variety of applications where graphs can be leveraged to perform data quality-enhancing tasks, such as outlier detection or virtual sensing among others. Before delving into specific use cases, let us roughly classify the tasks according to their scope:
\begin{itemize}
    \item \textit{Network level}: the outcome is obtained at the network (graph) level, i.e., the measurements at the different sensors and the graph topology are fed into a model to summarize the behavior of the collected measurements. For instance, detecting whether something is wrong within the network at a time instant, detecting events, or obtaining global metrics.
    \item \textit{Edge level}: in this case, the targets are the graph edges, e.g., link prediction. The goal is to predict or obtain a set of edges from the sensors' measurements. For instance, this approach can be used to estimate new connections between sensors, e.g., recently joined sensors, or the entire graph.
    \item \textit{Node (Sensor) level}: this is the most common case, where the target is the prediction of a feature at the sensor level, e.g., missing value imputation at different sensors using the sensor network measurements.
\end{itemize}

\begin{figure*}[!t]
    \centering
    \includegraphics[width=0.90\columnwidth]{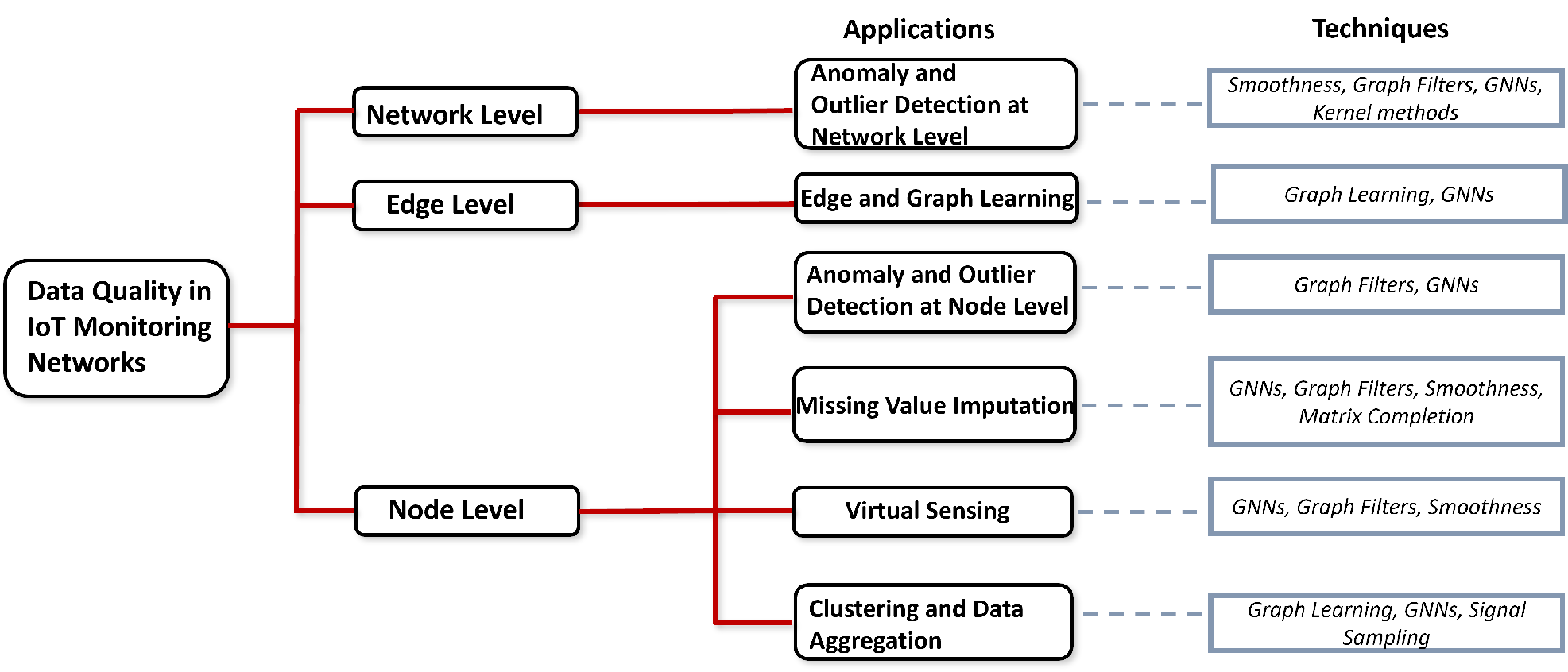}
    \caption{Proposed taxonomy for the different data quality-enhancing applications that can be carried out in a monitoring sensor network and the techniques that can be used.}
    \label{fig:taxonomy}
\end{figure*}

 Figure \ref{fig:tasks_levels} represents the different task levels. Moreover, Figure \ref{fig:taxonomy} presents the data quality-enhancement task taxonomy described in this section, as well as the theoretical aspects used for each one of the applications - more details are given in the following sections. At the \textit{network level}, we find tasks such as the \textit{detection of outliers and anomalies} which can be critical in assessing data consistency and coherence. At the \textit{edge level}, we find tasks such as \textit{edge prediction and graph learning} which might help in identifying correct graph structures and further applications to improve data accuracy. At last, at the \textit{node level}, we find applications such as \textit{outlier detection} to improve data consistency and coherence by identifying sensor measurement deviations, \textit{missing value imputation} to ensure data completeness, \textit{virtual sensing} to tackle different data quality dimensions such as accuracy or completeness, or \textit{clustering} which can be useful in grouping the data and improving data accuracy in further application. All these applications can be used to tackle data quality issues to produce continuous, complete, and accurate measurements \cite{ferrer2024data}.

\subsection{Network Level}
Let's define a network level task as finding the map $f_{\mathcal{G}}{:\;}\mathbb{R}^{N\times T}{\rightarrow\;}\mathbb{R}$ such that $y{=}f_{\mathcal{G}}(\mathbf{X})$ where $\mathbf{X}{\in}\mathbb{R}^{N\times T}$ is a graph signal ($T{=}1$) or a set of graph signals ($T{>}1$). The target variable is $y$, which can be a categorical variable (e.g., classification or outlier detection task) or a scalar value representing a metric for the graph.

\subsubsection{Anomaly and Outlier Detection at Network Level}
In the case of outlier and anomaly detection at the network level, we are interested in classifying a graph signal or a set of graph signals, i.e., $y{\in}\{1, \dots C\}$ where $C{\in}\mathbb{N}^+$ is the number of possible classes. For instance, the case of detecting whether the network measurements in a time instant correspond to an anomalous graph signal or not ($y{\in}\{0,1\}$). Hence, this approach does not provide localization capabilities, i.e., identifying the suspicious sensor or measurements that are causing the anomaly. Different ML approaches fit this setting; supervised ML, unsupervised learning, and semi-supervised learning. Supervised and semi-supervised models are limited in this field since labeled data about faulty sensors and abnormal measurements is difficult to obtain. The most common approach is the unsupervised learning setting where no labeled data is used and the data distribution together with the graph topology and a model are used. Ideally, data points, e.g., graph signals, that deviate from the distribution of the majority of observed graph signals are assumed to be anomalous.

GSP metrics such as total variation or graph signal smoothness have been used to identify anomalous graph signals, e.g., $S(\mathbf{S},\mathbf{x}){>}TH$, where $TH{\in}\mathbb{R}$ is a threshold value indicating the graph signals presenting discrepancies between connected sensors which are prone to be anomalous \cite{gopalakrishnan2019identification, chen2015signal}. Other GSP methods use the magnitude of the GDFT high-frequencies, the spectral graph wavelet transform, or the GDFT representation of a graph filter output to detect anomalous signals, e.g., $\|\hat{\mathbf{x}}_{\text{high}}\|{>}TH$, \cite{egilmez2014spectral, lewenfus2019use, xiao2020anomalous, sandryhaila2014discrete}. For instance, Xiao \textit{et al.} \cite{xiao2020anomalous} detected anomalies in a temperature monitoring IoT network by analyzing the frequency response of a graph filter.

Recently, Su \textit{et al.} \cite{su2024graph} have developed a graph-frequency domain filter which together with a Kalman filter is able to filter out anomalous graph signals in industrial pipe networks. Residual-based models have also been used to detect anomalies, where a graph signal reconstruction model, e.g., graph filter or GNN, is fitted with the data and large residuals' norm depicts a possible outlier, $\|\mathbf{x}-\mathbf{f}_{\mathcal{G}}(\mathbf{x})\|{>}TH$. Similarly, GNNs and GCNs have also been used as autoencoders to detect anomalous graph signals focusing on both signal reconstruction residuals and the reconstructed signal spectrum \cite{zhou2023robust, tang2022rethinking}. Moreover, spatiotemporal GNNs have been used to detect anomaly signals in multivariate time series, including anomalies in water monitoring sensor networks \cite{chen2021learning}. Lin \textit{et al.} \cite{lin2022deep} propose a context-aware graph autoencoder to detect anomalous air pollution events in an air quality monitoring network in Beijing. Feng \textit{et al.} \cite{feng2022full} propose a GAE based on graph convolutional layers to reconstruct the graph edges and graph signals and compute the anomaly score based on the loss function and the kernel density estimation of the distribution of scores during the training. Xu \textit{et al.} \cite{xu2021graph} propose an edge-based GCN to detect anomalous electricity consumption records in smart meter networks in a graph classification supervised manner.

Graph regularization can also be coupled with matrix factorization techniques, some examples include nonnegative matrix factorization or graph-regularized sparse representation for anomaly detection in other fields \cite{huang2018robust, li2021graph}.

\begin{figure}[!htp]
    \centering
    \includegraphics[width=0.75\columnwidth]{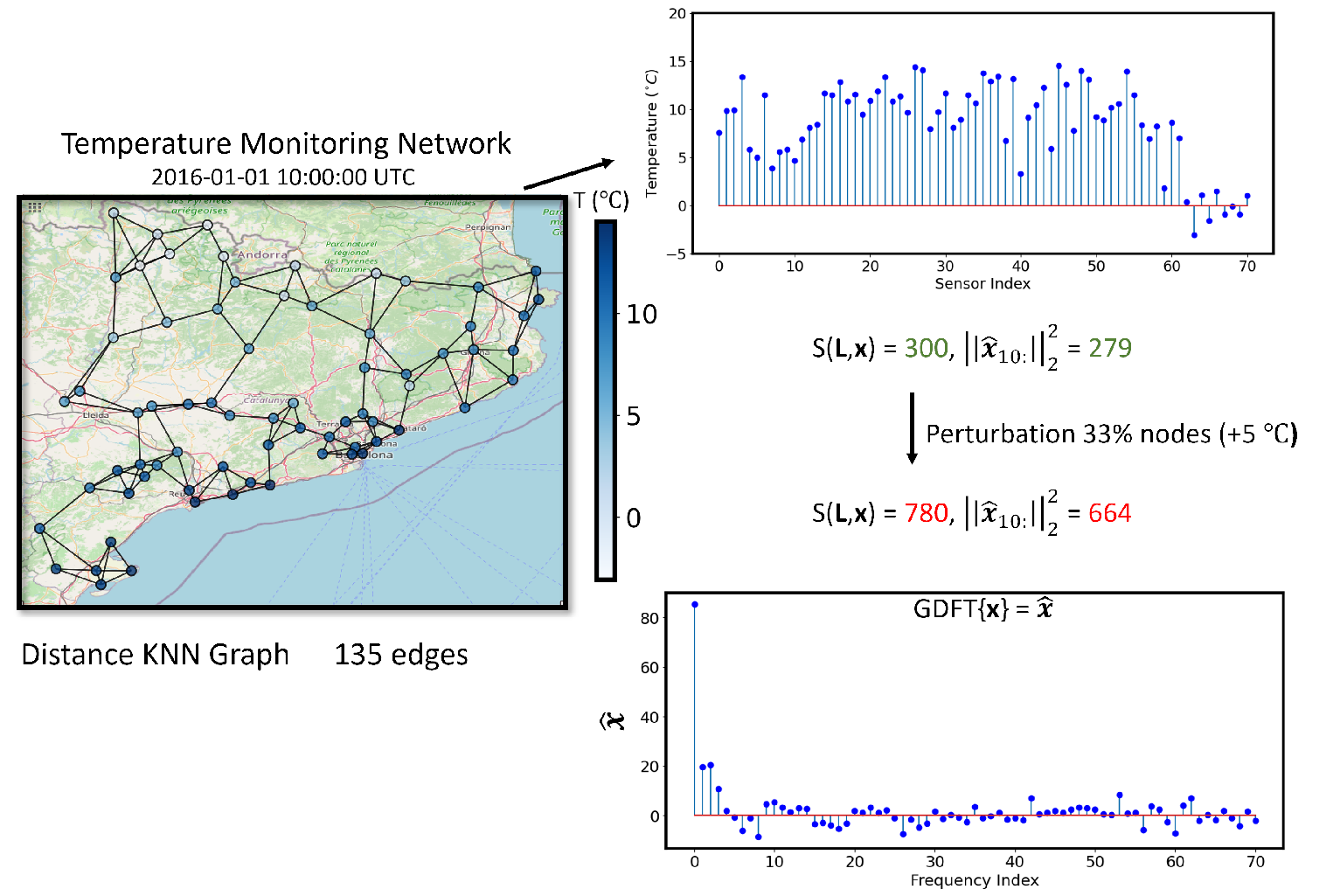}
    \caption{Example of a temperature monitoring network in the area of Catalonia, Spain, with a total of 71 nodes. The signal smoothness, $S(\mathbf{L}, \mathbf{x})$, and the norm of the GDFT coefficients associated to high-frequency components, $\|\hat{\mathbf{x}}_{10:}\|$, are used to detect anomalous measurements.}
    \label{fig:gdft_example}
\end{figure}

\textbf{Real-world use case}: Figure \ref{fig:gdft_example} shows an outlier detection example for a temperature monitoring network. This IoT monitoring network consists of 71 sensors deployed in the area of Catalonia, Spain. First, a distance-based K-nearest neighbors graph, section \ref{subsubsec:shift}, has been constructed, resulting in a Laplacian matrix $\mathbf{L}$ with 135 edges. Then a temperature signal has been taken at a specific time and the smoothness ($S(\mathbf{L},\mathbf{x})$=300) and magnitude of the higher-order coefficients of the GDFT ($\|\hat{\mathbf
{x}}\|_2^2$=279) have been computed. Then, 33\% of the nodes were perturbed with a shift of +5$^{\circ}C$ to simulate an anomalous signal. As it can be seen, both the smoothness and the magnitude of the higher-order GDFT coefficients can be used to detect the anomalous signal ($S(\mathbf{L},\mathbf{x})$=780 and $\|\hat{\mathbf
{x}}\|_2^2$=664). Other techniques could include the use of graph filters or GNNs and the subsequent evaluation of the norm of the residuals or the GDFT coefficients followed by a thresholding procedure. In similar real-world use cases, a model based on GDFT and a GCN were compared when applied to temperature sensor networks \cite{xiao2020anomalous}. Similarly, the GAE proposed by Feng \textit{et al.} \cite{feng2022full} was compared with VGAE in an industrial IoT network.

\begin{table*}[!htp]
\centering
\caption{Examples of anomaly and outlier detection graph-based models applied at the network level.}
\label{tab:network_OD}
\resizebox{1\textwidth}{!}{%
\begin{tabular}{@{}llllll@{}}
\toprule
 & & \textbf{Reference}  &  \textbf{Year}&\textbf{Method} & \textbf{Use Case} \\ \midrule
\multicolumn{1}{l|}{\multirow{8}{*}{\textbf{GSP}}} & \multirow{2}{*}{Smoothness} & Chen \textit{et al.} \cite{chen2015signal}        &          2015& \multicolumn{1}{|l}{\multirow{2}{*}{TV}}      &    Temperature sensor network      \\
 \multicolumn{1}{l|}{}& & Gopalakrishnan \textit{et al.} \cite{gopalakrishnan2019identification}         &      2019&     \multicolumn{1}{|l}{}  &  Delays US airports        \\
 \multicolumn{1}{l|}{} & & & & & \\
 \multicolumn{1}{l|}{} & \multirow{5}{*}{Graph Filter}& Egilmez \textit{et al.} \cite{egilmez2014spectral}        &      2014&High-Frequencies Graph Filter    &    Synthetic sensor network      \\
\multicolumn{1}{l|}{}  & & Lewenfus \textit{et al.} \cite{lewenfus2019use}      &          2019&Spectral Graph Wavelet Transform (SGWT)     &    \multicolumn{1}{|l}{\multirow{2}{*}{Temperature sensor network}}   \\
\multicolumn{1}{l|}{}  & & Sandryhaila \textit{et al.} \cite{sandryhaila2014discrete} &   2014&High-Frequencies GDFT & \multicolumn{1}{|l}{} \\
 \multicolumn{1}{l|}{} &   & Su \textit{et al.} \cite{su2024graph} &   2024&Graph-Frequency Domain Kalman Filter & Industrial Pipe Network \\
\multicolumn{1}{l|}{} &  & Xiao \textit{et al.} \cite{xiao2020anomalous} &  2020 & GDFT of Nonlinear Polynomial Graph Filter (NPGF) & Temperature sensor network  \\
  & &&&& \\
\multicolumn{1}{l|}{\multirow{4}{*}{\textbf{GNN}}}& \multirow{4}{*}{GCN \& GAE}& Chen \textit{et al.} \cite{chen2021learning} &   2021&GCN \& Attention Mechanism & Water sensor network \\
\multicolumn{1}{l|}{} & & Feng \textit{et al.}\cite{feng2022full} &  2022&GAE using GCN & Industrial IoT network \\

\multicolumn{1}{l|}{}  &   & Lin \textit{et al.} \cite{lin2022deep} &   2022&Context aware GAE & Air quality sensor network \\

\multicolumn{1}{l|}{}  & & Xu \textit{et al.} \cite{xu2021graph} &   2021&GCN & Smart Meter Network \\

   \bottomrule
\end{tabular}%
}
\end{table*}

All in all, residual-based models are used to detect anomalies and outliers in an unsupervised manner, the signal reconstruction can be implemented in different ways, e.g., convex optimization, graph filters, GNN, and the anomaly score is computed based on some norm or distance. Graph filters can provide a scalable, interpretable, and convex alternative to highly complex GNNs. Nevertheless, depending on the data availability, GNNs can provide superior performance due to their ability to model multivariate time series, including temporal components and complex surfaces. In any case, as we will see in section \ref{subsubsec:generalization}, GNNs and GSP tools can provide benefits in terms of model generalization and domain transferability, so easing the training of models and dealing with limited data scenarios. Table \ref{tab:network_OD} lists some graph-based models as well as their use cases.

\subsection{Edge Level}
Let's define an edge level task as discovering a set of graph edges $e {\subseteq} \mathcal{E}$ or learning a graph shift matrix $\mathbf{S}{\in}\mathbb{R}^{N\times N}$ given a set of measurements $\mathbf{X}{\in}\mathbb{R}^{N\times T}$ or some prior assumptions.
In this section, we place special emphasis on the use of graph learning, see section \ref{subsubsec:shift}, to learn the relationships between sensors in a network, thus revealing patterns and allowing the use of clustering or other graph-based techniques. As mentioned in section \ref{subsubsec:shift}, several graph learning techniques can be used for graph structure inferring \cite{dong2019learning, mateos2019connecting}. In most cases, the graph learning problem is formulated as an optimization problem where the graph shift matrix is restricted to the set of valid graph matrices. For instance, Jablonksi \cite{jablonski2017graph} applies a smoothness-based graph learning model to discover the relationships between reference stations of tropospheric ozone (O$_3$) to unveil the implicit relationships and apply clustering algorithms. Similarly, Mei \textit{et al.} \cite{mei2016signal} apply casual graph learning to infer the temperature diffusion pattern in the United States. Ferrer-Cid \textit{et al.} \cite{ferrer2021graph} discuss the use of different graph learning algorithms for representing air quality monitoring sensor networks. Likewise, Allka \textit{et al.} \cite{allka2024leveraging} perform graph learning over a sensor network to discover the relationships between the sensors and group the data to feed a neural network-based anomaly detector. Cini \textit{et al.} \cite{cini2023sparse} develop a sparse graph learning algorithm and applied it to an air quality index sensor network. Jiang \textit{et al.} \cite{jiang2021graph} propose an optimization problem to jointly perform graph learning and matrix completion over a temperature sensor network using a smoothness-promoting approach. Similarly, Liu \textit{et al.} \cite{liu2019graph} propose a graph learning model for time-varying graph signals that takes into account the spatiotemporal smoothness of the graph and provides good results for temperature sensor networks. Apart from techniques based on formulating the graph learning problem as an optimization problem using some assumptions such as signal smoothness, there are other approaches based on GNNs, e.g., VGAE, that have been used to infer the topology of the graph or estimate some of the links of the graph as well as to perform a specific task \cite{zhang2023vgae,dehmamy2019understanding}. Chen \textit{et al.} \cite{chen2021learning} use a learning policy to reduce the computational costs of learning the graph structure. In the context of precision agriculture, Vyas \textit{et al.} \cite{vyas2020dynamic} propose a soil moisture prediction task coupled with the identification of a dynamic graph structure.

In short, although there are approaches based on optimization and GNNs, for sensor networks it is still very common to establish the network topology from the geodesic distances between sensors. For instance, Do \textit{et al.} \cite{do2020graph} establish the graph adjacency matrix using the geodesic distances of sensors and locations for air quality sensor networks. Nonetheless, data-driven approaches can learn tailored graphs that better represent the implicit relationship between sensor nodes as well as graphs tailored to suit a specific task or application. These data-driven graphs can improve the performance of further applications, e.g., missing value imputation \cite{jiang2021graph, ferrer2022data}. Convex optimization problems can provide an informative alternative to GNNs that learn the graph representation for a specific application.

\subsection{Node Level}
Let's define a node-level task as finding the map $f_{\mathcal{G}}{:\;}\mathbb{R}^{N_1}{\rightarrow\;}\mathbb{R}^{N_2}$ such that $\mathbf{y}{=}f_{\mathcal{G}}(\mathbf{X})$ where $\mathbf{X}{\in}\mathbb{R}^{N_1\times T}$ is a graph signal ($T{=}1$) or a set of graph signals ($T{>}1$) and $\mathbf{y}$ is the target variable, which can be categorical. For instance, binary vector in an outlier detection task where each graph node can be identified as anomalous, or a scalar-valued vector for the prediction of graph signals at different nodes. Value $N_1$ denotes the number of observed nodes in the graph and value $N_2$ denotes the number of target nodes in the graph.

\subsubsection{Anomaly and Outlier Detection at Node Level}
\label{subsubsec:node_outlier}

Outlier and anomaly detection at the node level is especially interesting in IoT monitoring sensor networks since it localizes the sensor with faulty measurements and this can lead to further replacement or maintenance actions. This is known as outlier/anomaly localization. In this case, the output of the outlier detection process is $\mathbf{y}{\in}\{0, 1\}^N$, where each component $y_i$ indicates whether the $i$-th sensor is presenting an anomaly.

The most common approach for the outlier detection and localization task in monitoring sensor networks is the use of residual-based models. These models train a signal reconstruction model using data with normal patterns and then the reconstruction residuals are used to identify the outliers. Following the GSP framework, linear and nonlinear graph filters have been used in the detection of outliers, $|x_i - f_{\mathcal{G}_i}(\mathbf{x})|{>}TH$, for air quality and temperature sensor networks \cite{ferrer2022volterra, xiao2020nonlinear}. Some of these graph filters allow a distributed implementation, e.g., the distributed nonlinear polynomial graph filter (NPGF) \cite{xiao2021distributed}.

GNNs naturally adapt to this application, allowing the implementation of graph autoencoders using attention mechanisms, convolution, and pooling layers \cite{ren2023graph, deng2021graph}. This problem can also be seen as the detection of outliers in multivariate time series, where the time series of the sensor network measurements are fed to a graph convolutional adversarial network \cite{deng2022graph}. State-of-the-art techniques include the use of GNNs and graph convolutional autoencoders. For instance, Miele \textit{et al.} \cite{miele2022deep} propose a graph convolutional autoencoder to detect anomalies in wind monitoring sensor networks. Wu \textit{et al.} \cite{wu2023detecting} propose the use of spatiotemporal GCN and graph wavelet networks for the detection of malfunctioning fine particulate matter (PM$_{2.5}$) sensors in a large-scale sensor network, providing a centralized and a localized solution. Other works make use of spatiotemporal graphs and recurrent GCN and graph coarsening to detect anomalous sensors using sophisticated anomaly scores \cite{yang2022learning}. Darvishi \textit{et al.} \cite{darvishi2023deep} highlight the importance of outlier detection in IoT networks used in the development of digital twins, in particular, they use a deep recurrent GCN to create virtual sensors and use the residuals to detect sensor failures. Han \textit{et al.} \cite{han2022learning} propose a GCN that is fed by the representation learned by a sparse autoencoder. Residuals are standardized providing explicit anomaly localization and the maximum score is used as anomaly metric, showing good results for water sensor networks. Besides, other methods focus on obtaining a list of outlying nodes given a set of measurements, e.g., Francisquini \textit{et al.} \cite{francisquini2022community} develop a graph filter to detect consistently outlying nodes.  Table \ref{tab:node_level_OD} summarizes different graph-based models for outlier detection at the node level.

\begin{table*}[!t]
\centering
\caption{Examples of anomaly and outlier detection graph-based models applied at the node level.}
\label{tab:node_level_OD}
\resizebox{1\textwidth}{!}{%
\begin{tabular}{@{}llllll@{}}
\toprule
& & \textbf{Reference} &  \textbf{Year}&\textbf{Method} & \textbf{Use Case} \\ \midrule
\multicolumn{1}{l|}{\multirow{3}{*}{\textbf{GSP}}}&\multirow{3}{*}{Graph Filter}& Ferrer-Cid \textit{et al.} \cite{ferrer2022volterra}        &            2022&Volterra Graph Filter Outlier Detection (VGOD)      &    Air quality sensor network      \\
\multicolumn{1}{l|}{}&& Francisquini \textit{et al.} \cite{francisquini2022community} &   2022&Spectral Graph Filter Community-Based Detection & Water sensor network \\
\multicolumn{1}{l|}{} && Xiao \textit{et al.} \cite{xiao2020nonlinear}         &           2020& NPGF       &  Temperature sensor network        \\
& & & & & \\
\multicolumn{1}{l|}{\multirow{5}{*}{\textbf{GNN}}}&\multirow{5}{*}{GCN \& GAT}& Darvishi \textit{et al.} \cite{darvishi2023deep} &  2023 & Deep Recurrent GCN & \multicolumn{1}{|l}{\multirow{3}{*}{Water sensor network}} \\
\multicolumn{1}{l|}{}&& Deng \textit{et al.} \cite{deng2021graph} &  2021&Graph Learning \& GAT &  \multicolumn{1}{|l}{} \\
\multicolumn{1}{l|}{}&& Han \textit{et al.} \cite{han2022learning} &  2022&Sparse Autoencoder \& GCN & \multicolumn{1}{|l}{} \\
\multicolumn{1}{l|}{}&&  Wu \textit{et al.} \cite{wu2023detecting} &   2023&GCN \& Graph Wavelet Networks & Air quality sensor network \\
\multicolumn{1}{l|}{}&&  Yang \textit{et al.} \cite{yang2022learning} &    2022&Residual GCN \& Latent Spatial-Temporal Graph modeling (LSTGM) & Water Sensor Network \\
  \bottomrule
\end{tabular}%
}
\end{table*}

All in all, residual-based models are commonly used for anomaly and outlier detection in sensor networks since they allow for the computation of a metric per sensor, allowing the localization of the faulty sensor. The most common techniques include the use of graph filters, signal smoothness, and GNNs. The main advantage of GNNs is their ability to model complex and heterogeneous phenomena, as their architecture can include classical layers for multivariate cases, temporal components, or attention mechanisms. Nevertheless, graph filters provide an interpretable alternative that depending on the data can provide competitive performance, providing good distributed and generalizable approaches, section \ref{subsubsec:generalization}. Although not common, graph clustering techniques could also be employed to detect anomalous sensors in a sensor network given that they have been used successfully in other fields \cite{liu2021anomaly}.

\subsubsection{Missing Value Imputation}
Sensor networks are prone to losses, either due to sensor, communication or storage failures. Therefore, it is common for measurements to present losses, i.e., at time $t$, only a subset of sensors $\mathcal{M}{\subset}\mathcal{V}$ collect measures. Given its importance, missing value imputation has been a very active field of research, leading to the creation of different imputation models. This task is tackled at the node level since the main goal is to infer the graph signal components at different nodes, e.g., imputing the measurements of certain sensors in the network. The imputation of missing values in sensor networks is also referred to as matrix completion, graph signal reconstruction, or multivariate time series imputation or forecasting, among others.

Models based on matrix completion are transductive by nature, use a matrix of spatiotemporal measurements, $\mathbf{X}{\in}\mathbb{R}^{N\times T}$, with missing values, and their objective is to impute the missing entries of the matrix via the minimization of a specific matrix norm or regularization strategy. GSP-based matrix completion algorithms have been proposed where signal smoothness metrics, e.g., Sobolev smoothness, are optimized \cite{han2019large, deng2021graphvirtual, mondal2022recovery}. 
Other frameworks jointly learn the graph and complete the missing data via matrix completion using the $l_{2,2}$ and nuclear norms \cite{jiang2021graph}. Besides, matrix factorization techniques have also been used for missing value imputation by using graph kernels to model spatial relationships \cite{lei2022bayesian}. There also exist models that tackle the missing value imputation problem using graph signal reconstruction techniques based on the signal smoothness criterion, with applications in air quality sensor networks \cite{ferrer2022data, betancourt2023graph}. For instance, Qiu \textit{et al.} \cite{qiu2017time} impute PM$_{2.5}$ sensor measurements by optimizing the time-varying smoothness of graph signals.

\begin{table*}[!t]
\centering
\caption{Examples of graph-based missing value imputation methods at node level.}
\label{tab:node_MVI}
\resizebox{1\textwidth}{!}{%
\begin{tabular}{@{}llllll@{}}
\toprule
&&\textbf{Reference} &   Year&\textbf{Method} & \textbf{Use Case} \\ \midrule
\multicolumn{1}{l|}{\multirow{7}{*}{\textbf{GSP}}} & \multirow{7}{*}{Smoothness}&Betancourt \textit{et al.} \cite{betancourt2023graph}        &          2023&Random Forest \& Graph Signal Smoothing      &    \multicolumn{1}{|l}{\multirow{2}{*}{Air quality sensor network}}      \\
\multicolumn{1}{l|}{}& &Ferrer-Cid \textit{et al.} \cite{ferrer2022data}        &           2022&Graph Learning \& Signal Smoothness      &   \multicolumn{1}{|l}{}     \\
\multicolumn{1}{l|}{} & & Han \textit{et al.} \cite{han2019large}     &           2019&Matrix Completion with Signal Smoothness Regularization    &    Speed monitoring sensor network      \\
\multicolumn{1}{l|}{} & &Jiang \textit{et al.} \cite{jiang2021graph}        &      2021&Graph Learning \& Signal Smoothness Matrix Completion      &   Air quality sensor network      \\
\multicolumn{1}{l|}{}&& Lei \textit{et al.} \cite{lei2022bayesian}        &            2022&Kernel Signal Smoothness Matrix Factorization      &    Speed monitoring sensor network      \\
\multicolumn{1}{l|}{} &   & Mondal \textit{et al.} \cite{mondal2022recovery}         &          2022&Sobolev Norm Matrix Completion       &  Weather monitoring network        \\
\multicolumn{1}{l|}{} &  & Qiu \textit{et al.} \cite{qiu2017time} &  2017&Time-Varying Signal Smoothness & Air quality sensor network \\
 & & & & & \\

\multicolumn{1}{l|}{\multirow{9}{*}{\textbf{GNN}}}&\multirow{5}{*}{GCN}&Castro \textit{et al.} \cite{castro2023time} &  2023&Time GNN - GCN and Smoothness Regularization & \multicolumn{1}{|l}{\multirow{3}{*}{Air quality sensor network}} \\
\multicolumn{1}{l|}{}&&Chen \textit{et al.} \cite{chen2022adaptive}        &      2022&Adaptive GCN     &   \multicolumn{1}{|l}{}     \\
  \multicolumn{1}{l|}{}&& Spinelli \textit{et al.} \cite{spinelli2020missing}        &      2020&Adversarial Graph Denoising Autoencoder     &     \multicolumn{1}{|l}{}   \\
   \multicolumn{1}{l|}{}& & Wu \textit{et al.} \cite{wu2021inductive}        &     2021&  Inductive GNN Kriging (IGNNK)     &   Precipitation monitoring network      \\
\multicolumn{1}{l|}{}&& Zhang \textit{et al.} \cite{zhang2023data}        &     2023&Spatiotemporal Variational Autoencoder    &   Weather monitoring network      \\

\multicolumn{1}{l|}{}&& & & & \\

\multicolumn{1}{l|}{}&\multirow{3}{*}{Other GNN}&Cini \textit{et al.} \cite{cini2022filling}        &    2022& GRIN   &   Smart meter network      \\
\multicolumn{1}{l|}{}&  & Marisca \textit{et al.} \cite{marisca2022learning}        &    2022&Sparse Spatiotemporal Attention GNN    &   Air quality sensor network      \\
  \multicolumn{1}{l|}{}&& Zhang \textit{et al.} \cite{zhang2023missing}        &       2023& GLPN     &   Solar power sensor network      \\

\bottomrule
\end{tabular}%
}
\end{table*}

In terms of GNNs, these have been widely used for imputing multivariate time series, which can be interpreted as a set of graph signals collected by sensor networks. The most common approach for matrix completion using GNNs is to train a GNN in the form of an autoencoder so that the reconstruction error over the non-missing entries of the matrix of observations is minimized \cite{spinelli2020missing}. It is worth noting that ground-truth values for missing entries are not available, therefore, loss functions can be computed over observed measurements or surrogate objective functions, e.g., artificially introduced missings or forecasting \cite{cini2022filling}. Examples of GNN architectures used include gated GNNs \cite{liu2020handling}, GCNs \cite{chen2022adaptive, zhang2023missing} or adversarially-trained GCNs \cite{spinelli2020missing}. For instance, Chen \textit{et al.} \cite{chen2022adaptive} develop a bidirectional recurrent GNN based on graph convolutional layers to perform matrix completion and online imputations in environmental data sets. Similarly, Zhang \textit{et al.} \cite{zhang2023data} propose a spatiotemporal variational autoencoder composed of graph convolution layers and recurrent units to impute missing values of multiple pollutants in environmental data sets. Zhang \textit{et al.} \cite{zhang2023missing} introduce the graph Laplacian pyramid network (GLPN) \cite{zhang2023missing} which proposes a signal's Dirichlet energy maintenance step to refine the imputation. Other GNNs have been used for missing value imputation like graph recurrent imputation network (GRIN) \cite{cini2022filling}, inductive graph neural networks that provide generalizable architectures to graphs of different topologies \cite{wu2021inductive}, or GNNs with sparse spatiotemporal attention mechanisms \cite{marisca2022learning}. Table \ref{tab:node_MVI} summarizes the different missing value imputation methods described above. Even though convex models based on GSP have been proven useful for missing value imputation, GNNs can provide superior performance for large data sets given their ability to model highly complex functions.

\begin{figure}[!h]
    \centering
    \includegraphics[width=0.75\columnwidth]{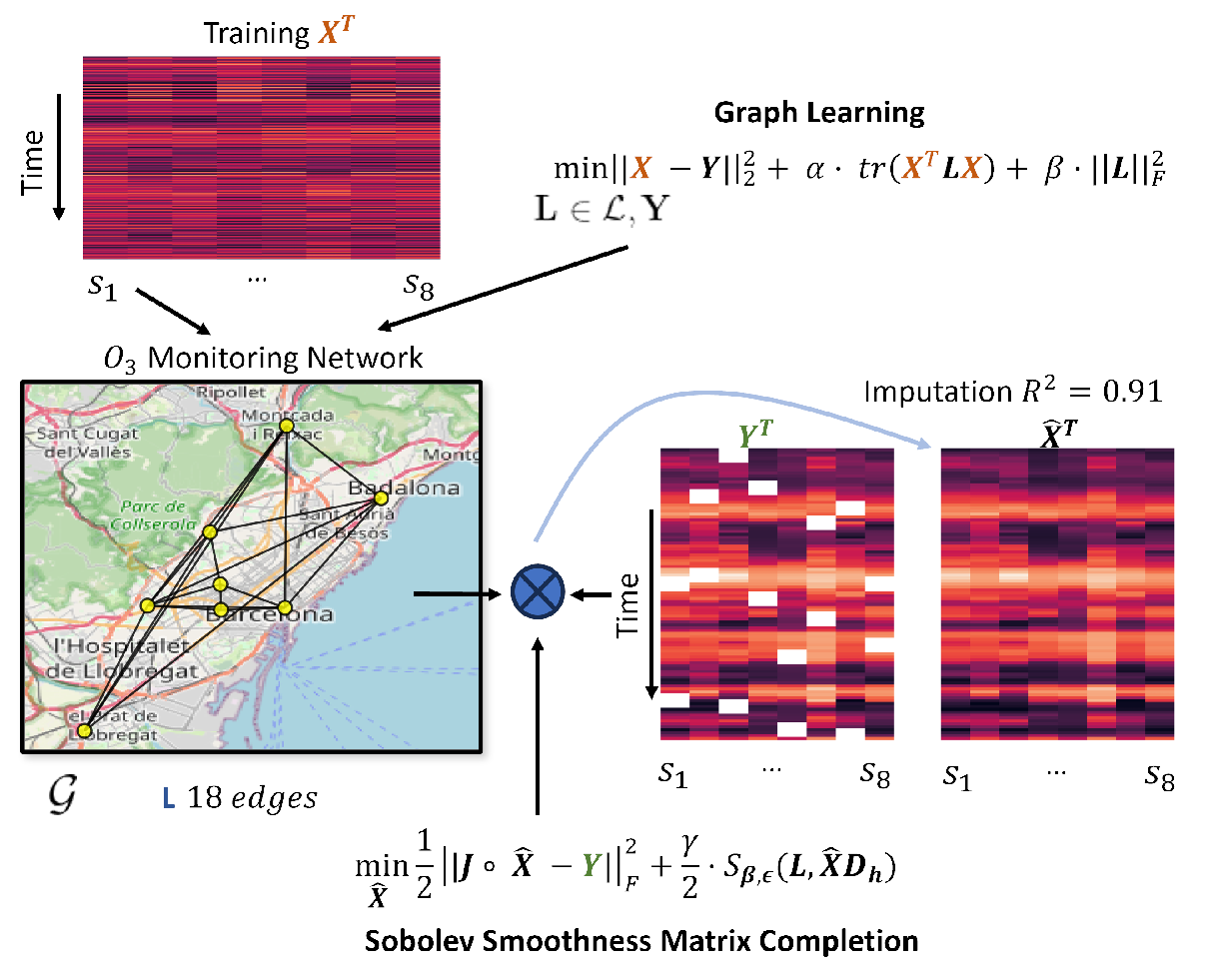}
    \caption{Example of a missing value imputation scenario in a $O_3$ monitoring network in the Barcelona area, Spain. A matrix completion model based on the Sobolev smoothness and a learned graph is applied to fill the data gaps, resulting in an imputation R$^2$ of 0.91.}
    \label{fig:mvi_num_example}
\end{figure}

\textbf{Real-world use case}: Figure \ref{fig:mvi_num_example} shows a missing value imputation example for an air quality monitoring network. The IoT monitoring platform measures tropospheric ozone (O$_3$) at 8 sites in the Barcelona area, Spain. Firstly, a data-driven graph (Laplacian matrix $\mathbf{L}$) is learned from the data matrix $\mathbf{X}{\in}\mathbb{R}^{N\times T}$, where $N$ is the number of sensors and $T$ the number of available measurements, using the methodology described in \cite{dong2016learning}. Then, matrix $\mathbf{Y}$ represents the collected network measures for a given time period, where there are some data gaps or missing values. To impute the missing values, a Sobolev smoothness matrix completion approach is used \cite{mondal2022recovery}. The imputation results in an R$^2$=0.91. Other graph-based missing value imputation approaches could be utilized, e.g., other signal smoothness-based criteria or GCNs making use of the learned shift matrix $\mathbf{L}$. Online variants or small data batches could be leveraged to impute missing values in real time. In similar real-world use cases, the adaptive GCN proposed by Chen \textit{et al.} \cite{chen2022adaptive} was compared with the GRIN for the imputation of PM$_{2.5}$, radiation, and humidity measurements in a network of over 500 sensors.

All in all, graph topologies provide an explicit approach to modeling the spatiotemporal relationships between sensors, which are exploited by graph-based models. It is worth mentioning that the reconstruction of a sensor signal, i.e., the creation of a virtual sensor for that sensor, can be used to impute measurements in case of losses and vice versa, a scenario seen in the next section \ref{subsubsec:vs}. Most effective techniques include matrix completion methods that pose a graph smoothness-based optimization method to complete graph signals. These techniques may face scalability challenges in large-scale sensor networks, e.g., advanced optimization techniques like the alternating direction method of multipliers (ADMM) have been used to ease scalability problems (see section \ref{subsec:challenges}). As mentioned in the previous section, the advantage of GNNs is that they can include a large variety of components to deal with highly complex functions, e.g., spatiotemporal components or attention mechanisms.

\subsubsection{Virtual Sensing}
\label{subsubsec:vs}

Virtual sensing or the creation of virtual sensors (also known as soft sensors) is a widespread application in monitoring sensor networks, where estimates for a given phenomenon and location are produced without having a physical sensor, this is the major difference with the missing value imputation task. Common application cases include the creation of virtual sensors to impute missing values, replace readings from faulty sensors, or detect faulty sensors. Therefore, we are estimating the measurements in a given node $x_{vs}$ using the measurements of the other graph nodes $\mathbf{x}_{\mathcal{M}}$. This can be tackled as a graph signal reconstruction or recovery task at a node where there is no underlying physical sensor. More generally, the creation of virtual sensors can be viewed as the signal reconstruction of a node in a graph, the estimation of an unobserved parameter, or the forecasting of multivariate time series. Another example of the application of virtual sensors is the case of dynamic sensor networks where new nodes representing physical sensors and virtual sensors can join or leave the network at any time. In the ML field, this case is also known as out-of-sample imputation while missing value imputation corresponds to in-sample imputation \cite{jin2023survey}.

\begin{table*}[!t]
\centering
\caption{Examples of virtual sensor applications using graph-based models at the node level.}
\label{tab:node_virtual_sensing}
\resizebox{1\textwidth}{!}{%
\begin{tabular}{@{}llllll@{}}
\toprule
& & \textbf{Reference} &  \textbf{Year}&\textbf{Method} & \textbf{Use Case} \\ \midrule
\multicolumn{1}{l|}{\multirow{4}{*}{\textbf{GSP}}}& \multirow{2}{*}{Smoothness}&   Chen \textit{et al.} \cite{chen2015signal}        &    2015& TV   &    \multicolumn{1}{|l}{\multirow{2}{*}{Air quality sensor network}}      \\
 \multicolumn{1}{l|}{}&& Ferrer-Cid \textit{et al.} \cite{ferrer2022data}        &      2022&Graph Learning \& Signal Smoothness      &       \multicolumn{1}{|l}{}   \\
\multicolumn{1}{l|}{}& & & & & \\
\multicolumn{1}{l|}{} &Graph Filter  & Ferrer-Cid \textit{et al.} \cite{ferrer2024virtual}      &  2024 & Volterra-Based Graph Filter              & Environmental monitoring network \\
& & & & & \\
\multicolumn{2}{l}{Matrix Factorization}&   Lei \textit{et al.} \cite{lei2022bayesian}        &            2022&Bayesian Graph Kernelized Matrix Factorization      &    Speed monitoring sensor network      \\
& & & & & \\

\multicolumn{1}{l|}{\multirow{15}{*}{\textbf{GNN}}}& \multirow{8}{*}{GCN}& Castro \textit{et al.} \cite{castro2023time} &  2023&Time GNN - GCN and Smoothness Regularization & Air quality sensor network \\
\multicolumn{1}{l|}{}& &  Custodio \textit{et al.} \cite{custodio2024comparing} &  2024&Spectral Temporal GNN & Precision Agriculture \\
\multicolumn{1}{l|}{}&& Darvishi \textit{et al.} \cite{darvishi2023deep} &  2023 & Deep Recurrent GCN & Water sensor network \\
\multicolumn{1}{l|}{}&&  Do \textit{et al.} \cite{do2019matrix}        &     2019&  \multicolumn{1}{|l}{\multirow{2}{*}{Variational GAE Matrix Completion}}     &   \multicolumn{1}{|l}{\multirow{2}{*}{Air quality sensor network}}      \\
\multicolumn{1}{l|}{}& & Do \textit{et al.} \cite{do2020graph}        &    2020&  \multicolumn{1}{|l}{}    &  \multicolumn{1}{|l}{}       \\
\multicolumn{1}{l|}{}&  & Jia \textit{et al.} \cite{jia2023graph} &   2023&GCN \& Temporal Convolution & Industrial process network \\
 \multicolumn{1}{l|}{}& & Wu \textit{et al.} \cite{wu2021inductive}        &        2021& IGNNK   &   Precipitation monitoring network      \\
\multicolumn{1}{l|}{}&&  Zhao \textit{et al.} \cite{zhao2024graph} & 2024 & Heterogeneous Temporal GNN & Industrial sensor network \\

 \multicolumn{1}{l|}{}& & & & & \\

\multicolumn{1}{l|}{}&\multirow{6}{*}{Other GNN}& Calo \textit{et al.} \cite{CALO2024108191} &  2024&Message Passing Algorithm & Air quality sensor network \\
\multicolumn{1}{l|}{}&&  Cini \textit{et al.} \cite{cini2022filling}        &     2022& GRIN   &   Smart meter sensor network      \\
 \multicolumn{1}{l|}{}& & Niresi \textit{et al.} \cite{niresi2023spatial} &  2023&GAT \& LSTM & Air quality sensor network \\
\multicolumn{1}{l|}{}&  & Vyas \textit{et al.} \cite{vyas2020dynamic} &  2022&Recurrent GNN & Precision Agriculture \\

\multicolumn{1}{l|}{}&& Yan \textit{et al.} \cite{yan2022virtual}        &    2022& GAN \&  GAT   &   Industrial IoT network      \\
\multicolumn{1}{l|}{}&& Zhang \textit{et al.} \cite{zhang2023missing}        &   2023& GLPN &   Solar power sensor network      \\

  \bottomrule
\end{tabular}%
}
\end{table*}

Virtual sensors have been created using graph signal reconstruction and GNNs methods \cite{chen2015signal, castro2023time, jin2023survey}. For instance, Ferrer-Cid \textit{et al.} \cite{ferrer2022data} use the signal smoothness criterion to estimate virtual sensors for O$_3$ sensors. In fact, graph signal reconstruction models based on GSP can be used to implement virtual sensors given that the measurements at non-sampled vertices (sensors) are reconstructed using the network measurements \cite{castro2023time,qiu2017time}. Lei \textit{et al.} \cite{lei2022bayesian} propose a Bayesian kernelized matrix factorization model to approach the missing value imputation problem as well as the virtual sensor problem, establishing a connection with Kriging where it is intended to predict in a location where there is no sensor. Graph kernels are used to model spatial relationships.

GNNs have been widely used for virtual sensing applications. Yan \textit{et al.} \cite{yan2022virtual} make use of a graph attention network to create virtual sensors for rocket-mounted sensors, which are later used to detect faulty sensors. Cini \textit{et al.} \cite{cini2022filling} also make use of the GRIN to create virtual sensors for an air quality sensor network. Similarly, Wu \textit{et al.} \cite{wu2021inductive} introduce the inductive graph neural network Kriging (IGNNK) capable of generalizing to unobserved nodes, therefore, creating virtual sensors. Moreover, matrix completion using variational graph autoencoders has been carried out to infer air pollutants at a fine-grain scale, thus, creating virtual sensors for unmonitored locations \cite{do2019matrix, do2020graph}. Following a similar approach, Calo \textit{et al.} \cite{CALO2024108191} introduce a spatial air signal prediction method to estimate the pollution at unobserved locations (street level) using a mean aggregation message passing algorithm in conjunction with a GNN. Indeed, this approach can be seen as obtaining virtual sensors for those locations where the signal is predicted. In terms of phenomena estimation, which can be seen as the implementation of a virtual sensor, Jia \textit{et al.} \cite{jia2023graph} develop a combination of GCN with a temporal convolutional network to estimate the quality of an industrial process. In another interesting example, Niresi \textit{et al.} \cite{niresi2023spatial} propose a spatiotemporal data fusion GAT to perform air pollution sensor calibration, which can be viewed as obtaining a virtual sensor that fuses measurements from different sensors. In the context of precision agriculture, Vyas \textit{et al.} \cite{vyas2020dynamic} use a recurrent GNN for soil moisture forecasting from other related environmental parameters. Similarly, Custodio \textit{et al.} \cite{custodio2024comparing} evaluate the performance of the spectral temporal GNN, which combines the GDFT and classic layers such as one-dimensional convolutions, to perform soil moisture forecasting from temperature and weather parameters. Zhao \textit{et al.} \cite{zhao2024graph} introduce the heterogeneous temporal GNN, which can capture the dynamics of different sensors and include the effect of exogenous variables. Darvishi \textit{et al.} \cite{darvishi2023deep} use GCN-implemented virtual sensors to detect outliers and replace erroneous sensor readings with those of the virtual sensor if necessary, with a focus on networks feeding digital twins.

This field is continuously growing, as new graph signal reconstruction techniques, matrix completion techniques, and generalization frameworks are appearing and can be used to create virtual sensors in the context of IoT monitoring sensor networks. Table \ref{tab:node_virtual_sensing} lists different state-of-the-art models that have been used for the estimation of virtual sensors. As it can be noted, most virtual sensing models are based on GNNs, given their ability to accommodate highly complex spatiotemporal relationships \cite{dong2023graph}. Virtual sensors are widely used, whether for outlier detection, sensor replacement, or even parameter estimation at unmonitored sites. The performance of the models depends very much on the nature of the network, whether it is heterogeneous (sensors measuring different phenomena), the amount of data available, or even the smoothness of the measurements. GNNs are the most widely used technique for virtual sensing, transferability and domain generalization properties can be key to their successful application in scenarios where data may be limited. In short, virtual sensing is a key application that allows for uninterrupted measurements and detection of possible anomalies or deviations, which are very important features in scenarios where the data feed digital twins.

\subsubsection{Clustering and Data Aggregation}
Node clustering or data aggregation and compression are tasks that are usually carried out in wireless sensor networks. In fact, clustering can be used to group nodes for communication purposes, data compression and aggregation to save energy, or for applying divide-and-conquer-like algorithms. Therefore, the task of discovering $K$ clusters can be seen as finding the map $\mathbf{y}{=}f_{\mathcal{G}}(\mathbf{X})$, where $\mathbf{X}{\in}\mathbb{R}^{N\times T}$ is a set of network measurements and $\mathbf{y}{=}\{1,\dots,K\}^N$ classifies sensors into clusters. In the context of node clustering, deriving from graph spectral theory, the Laplacian eigenvectors have been used for clustering air pollution reference stations \cite{jablonski2017graph}. Ji \textit{et al.} \cite{ji2021smoothness} propose a node-level smoothness criterion combined with a self-supervised GCN to perform graph clustering, although not applied, this work gives an idea of possible applications in monitoring sensor networks. Indeed, the Laplacian fielder vector and Laplacian spectral graph theory have been widely used for the clustering of network nodes \cite{ortega2018graph, muniraju2017location,thangaramya2017energy, riahi2017using}.

In the context of data aggregation and compression graph-based techniques have been developed \cite{manuel2022data}. Sensor sampling schedulers have been designed using graph signal reconstruction techniques, so that the best subset of sensors is found to minimize the signal reconstruction error at the unmonitored locations, reducing the overall network energy consumption. Signal reconstruction techniques based on the GDFT have been used for the sensor sampling problem \cite{rusu2017node}. Recently, a graph sampling theory-based model has been developed for dynamic sensor placement, i.e., sensors are not static but can change positions \cite{nomura2024dynamic}.

\section{Challenges, Limitations, and Emerging Trends}
\label{sec:future}
In this section, we discuss some relevant challenges and limitations that have arisen, as well as identify emerging trends in signal processing over graphs. However, we remark that the applications reviewed, such as virtual sensing, outlier detection, or missing value imputation, are very active research fields.

\subsection{Challenges and Limitations}
\label{subsec:challenges}
In this section, we describe some of the limitations and challenges that graph-based models face in modeling and improving IoT network data quality.

\textbf{Large-scale IoT monitoring networks}: one of the main challenges is the modeling of large-scale sensor networks. In this scenario, a graph can have thousands of nodes, so not only can the training of data quality enhancement models be complex, but the construction of the graph itself can be challenging. In the realm of GSP, linear algebra features, such as vectorization or parallelization of operations, have been used to define operations such as GDFT in high-dimensional scenarios \cite{bigdata2014}. In this line, specialized optimization techniques, such as the alternating direction method of multipliers (ADMM) have been used to solve large-scale modeling GSP-based problems \cite{han2019large,dong2016learning}. Moreover, GNNs provide operations, such as graph sampling or pooling, that allow efficient learning in high-dimensional graphs. In this regard, two fields have emerged that allow for optimizing training in large-scale networks, such as distributed training and the generalization and transference of models to larger networks (section \ref{subsubsec:generalization}).

\textbf{Data availability}: is one of the main requirements for training complex data-driven models. This aspect is also relevant in scenarios where IoT platforms are modeled with little data, but similar IoT data is available in other environments. Therefore, several research lines have been developed that study the transference of pre-trained models and the generalization of trained models on small networks to larger networks \cite{zhou2020graph} (section \ref{subsubsec:generalization}).

\textbf{Heterogeneity and dynamic graphs}: Another aspect to consider is the complexity and heterogeneity of IoT networks. Although we have focused on static IoT monitoring networks in this survey, more advanced networks, such as mobile networks, not only have a variable number of nodes (since a node can enter or leave the network at any time), but also the relationships between nodes can change. Therefore, mobile monitoring networks would require the use of time-varying graphs \cite{yan2024signal}. Another aspect that has attracted attention is the assumption of smoothness and homophily used in graph-based models, where connected nodes tend to be similar. This assumption can be compromised in networks where sensors that measure different phenomena coexist or where there are different types of nodes. Recent studies analyze the performance of graph-based models such as GNNs in applications under heterophily \cite{platonov2023critical}.

\textbf{Security}: the security of graph models has been one of the most overlooked topics. As previously observed, the robustness of GNNs can be compromised by adversarial attacks \cite{zhou2020graph}. Therefore, different attacks have been analyzed and the robustness of GNNs and GSP models has been the subject of study during recent years \cite{liu2024towards,sun2022adversarial,hosseini2024comprehensive}. This topic is covered in section \ref{subsubsec:security}. Another related aspect is data privacy and ownership, with federated learning approaches being explored to ensure privacy in IoT applications (section \ref{subsubsec:federated}).

\subsection{Emerging Trends}
\label{subsubsec:trends}
\begin{figure}[!h]
    \centering
    \includegraphics[width=0.85\textwidth]{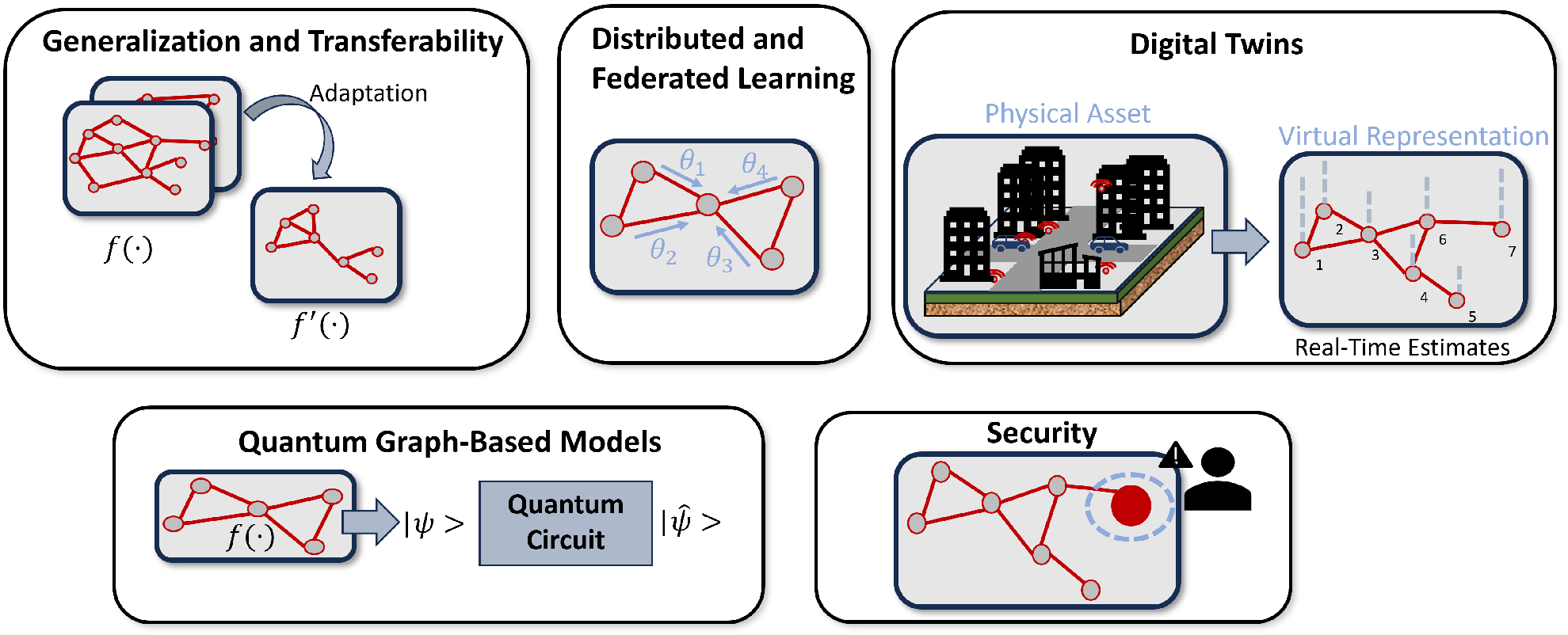}
    \caption{A number of emerging trends that take advantage of important features offered by graph-based models.}
    \label{fig:emerging}
\end{figure}

In this section, we review some emerging trends in graph-based models for IoT monitoring networks. We show some of the trends that arise in response to some of the challenges and limitations presented in the previous section. Figure \ref{fig:emerging} represents the different emerging trends reviewed. Table \ref{tab:challenges} describes the graph features leveraged in the emerging trends explained in this section \ref{subsubsec:trends}.

\begin{table}[!h]
\small
\centering
\caption{Summarization of the different emerging trends highlighted in the paper.}
\label{tab:challenges}
\resizebox{0.78\columnwidth}{!}{%
\begin{tabular}{p{50mm}p{75mm}}
\toprule
\footnotesize
\textbf{Emerging Trends }           & \textbf{Graph Features} \\ \midrule
\textbf{(1)} Generalization and Transferability &      Graph-based models provide intradomain transferability and adaptation properties.          \\
\textbf{(2)}  Distributed and Federated Learning    & Most of graph filters allow a distributed implementation.          \\
\textbf{(3)} Digital Twins                         &     Graph-based models can be used to produce estimates for an asset represented by a graph.\\
\textbf{(4)} Quantum Graph-Based Models            &     GNNs can be translated into quantum-inspired algorithms.           \\
\textbf{(5)} Security           &     Graph-based models need to provide robustness against cyberattacks.           \\
\bottomrule
\end{tabular}%
}
\end{table}

\subsubsection{Generalization, Transferability, and Transfer Learning}
\label{subsubsec:generalization}
One of the relevant aspects of graph-based models and more specifically of GNNs is their generalization and transferability, which is a very active research field. When a GNN is trained, the mechanism of message passing or propagation of measurements between nodes is learned, instead of specific relationships between nodes, therefore, models trained on similar networks are prone to be transferable. The generalization and transferability capabilities of GNNs have been highlighted, where ideally models trained on small networks could be transferred to other networks, possibly of larger scale \cite{ruiz2023transferability, ruiz2020graphon, levie2021transferability, maskey2023transferability}. The main assumption required is that different networks measure the same phenomenon so that a GNN trained on one network has a similar effect on another network measuring the same phenomenon. The transferability of GNNs is partly inherited from the transferability of graph filters, more specifically spectral graph filters \cite{levie2019transferability}. For instance, Ruiz \textit{et al.} \cite{ruiz2023transferability} analyze the transferability of graph filters and GNNs, placing special emphasis on the transfer errors depending on the size of the network, where the transfer error is seen to decrease when increasing the original graph size. These ideas are along the lines of transfer learning, where a model, e.g., graph filter or GNN, is pre-trained and then optimized over a new network \cite{zhu2021transfer, gritsenko2023graph, liao2022deep}. Zhu \textit{et al.} \cite{zhu2021transfer} develop the ego-graph information maximization to transfer GNNs between graphs. Similarly, graph-based meta-learning techniques can be developed to tackle few-shot learning problems, where a GNN can be trained on a few samples from different data sets and can be applied to a few-shot data set \cite{zhou2019metagnn, mandal2022metalearning, spinelli2022meta}.

We believe that these approaches will allow the development of more efficient graph-based techniques to carry out different applications in IoT monitoring sensor networks, e.g., missing value imputation or virtual sensing. For instance, in air quality, governmental monitoring networks collect measurements without interruption, so these measurements can be used to create models that can then be applied to low-cost sensor networks, which have no (or little) historical data to train such models.

\subsubsection{Distributed and Federated Learning}
\label{subsubsec:federated}
One of the relevant aspects of the GSP and graph filters is the fact that by explicitly using the graph shift matrix in their operations, which represents the relationship between nodes, they can be implemented in a distributed manner \cite{li2021privacy}. For instance, Xiao \textit{et al.} \cite{xiao2021distributed} propose the graph Volterra series filter counterpart, which explicitly computes its output based on shifted versions of a graph signal and allows for a distributed approach. Other works have focused on improving the distributed application of graph filters by reducing the communication and computational resources required \cite{coutino2019advances}. The training and inference of distributed models in an IoT environment imply the adoption of an edge computing paradigm that needs to be aware of the resources available in the IoT nodes \cite{murshed2021machine}. As in the case of graph filters, GNNs have been the focus of research in their distributed application, given the massive size of networks and data that are used to train such neural networks. Thus, it has been actively investigated how to distribute the training and applications of GNNs in clusters of machines trying to optimize the imbalance in communications and workload, in a parallel fashion \cite{shao2024distributed, besta2022parallel, zheng2020distdgl}. For instance, Protogerou \textit{et al.} \cite{protogerou2021graph} propose a distributed application of a GNN using multi-agents that work cooperatively to perform anomaly detection tasks.

Just as distributed learning has been applied in graph-based applications, the development of federated learning schemes has also been recently applied in graph-based models \cite{liu2022federated, xie2021graph, mei2019sgnn, rasti2022graph}. In this case, different models are trained independently at each node using the data set available at each node, then these independent models are combined to obtain the final model. This approach is used for applications that require preserving data privacy and ownership, as well as security.

\subsubsection{Digital Twins}
Digital twins are gaining popularity in the academic and industrial environment, and this technology is increasingly intended to be transferred to other fields, e.g., smart cities, e-health, precision agriculture, etc \cite{thelen2022comprehensive}. A key component of digital twins is sensing, which complements the physical representation of an entity and provides feedback to improve an underlying physical model. Therefore, sensor networks play a key role in this paradigm, especially the quality of the data they provide.

Graph-based models, and more precisely GNNs, have been used to represent physical entities, therefore, producing digital twins. For instance, in the field of computer communications, GNNs have been used to predict network performance and behavior, being able to detect abnormal conditions \cite{yu2023digital, wang2020graph}. Ferriol-Galmés \textit{et al.} \cite{ferriol2022building} propose a GNN-based network digital twin to predict network performance and service level agreement metrics.

Applications have also been found in monitoring systems, where digital twins can help to detect different events and trigger the necessary actions. For instance, Roudbari \textit{et al.} \cite{roudbari2024data} use a GCRN in conjunction with a data-driven graph to predict water levels in a water monitoring network and feed a digital twin to assess potential flooding. Similarly, Sui \textit{et al.} \cite{sui2022graph} use a graph-based digital twin to model a power system, in the so-called Internet of Energy (IoE), and apply a GCN to detect possible faults and estimate the stability of the system. In the realm of smart healthcare, GNNs have been used to estimate and forecast patients' conditions \cite{barbiero2021graph}.

In general, IoT data can be fed into graph-based models that accurately represent physical entities and relationships between monitoring sites to generate useful insights that can later be used to support decision-making processes, perform early warning, predictive maintenance, or anomaly detection. As noted in related publications, graph-based models can represent and provide real-time estimates for an entity or asset. These estimates can be used to detect possible events and failures, or even predict future conditions. Furthermore, as seen in previous sections, graph-powered models are not only effective in estimating digital twins but can also provide solutions to improve the operation of sensor networks that feed digital twins, e.g., through malfunctioning sensor identification and replacement \cite{darvishi2023deep}. Therefore, much recent work has focused on improving the data quality and operability of the sensor networks that feed these digital twins.

\subsubsection{Quantum Graph-Based Models}
Recently with the rise of quantum computing, many quantum algorithms, or so-called quantum-inspired algorithms, have appeared that bring quantum mechanics fundamentals to ML algorithms to provide new properties, e.g., compact parameter representation, and faster training. Thus, much research has focused on developing the quantum analog of known techniques. This phenomenon has also reached graph-based models, particularly in the development of quantum GNNs. For instance, Verdon \textit{et al.} \cite{verdon2019quantum} introduce the quantum GNNs to represent quantum processes that possess a graph structure. Quantum recurrent GNNs and quantum convolutional GNNs have also been proposed, however, these specific architectures are tailored to a given application, such as graph classification or others \cite{zheng2021quantum, choi2021tutorial}. Lately, quantum graph transformers have been proposed, which aim to translate and compute graph transformers in quantum computing primitives and processors \cite{kollias2023quantum}. In short, we observe how graph-based models such as GNN are translated into quantum-inspired models, i.e., models that bring ideas from quantum computing to improve their classical analog, and other models that aim to use a pure quantum approach with quantum processors. However, we see how a small number of applications have been tackled with quantum graph-based models, e.g., graph classification, and we believe that in the future more applications and tasks will benefit from quantum-inspired and pure-quantum approaches.

\subsubsection{Security}
\label{subsubsec:security}
Model security and robustness have been some of the most overlooked aspects in the development of graph-based models \cite{zhou2020graph}. This is particularly interesting in an era where IoT data is being used in decision-making processes or, as seen above, in digital twin applications. Among the most known attacks are data poisoning, noise injection, or adversarial attacks on GNNs. In the case of GNNs, adversarial attacks affecting the robustness of the methods have been described. In particular, these attacks may include not only the modification of values but also the modification of the graph structure \cite{guan2024graph,sun2022adversarial,liu2024towards}. In terms of defense, just as attackers can use adversarial samples, adversarial training can be introduced into the GNN training methodology to provide robustness against possible attacks \cite{guan2024graph, sun2022adversarial}. Another defense method is to detect possible modifications of samples, nodes, or edges of the graph, and eliminate possible malicious elements. Similarly, in the context of GSP, GDFT-based techniques have been used to introduce structural adversarial samples, and various GSP methods, such as GDFT or smoothness, have been shown to be useful in detecting maliciously injected data \cite{liu2023point, hosseini2024comprehensive}.

All in all, the models used must be robust to ensure the integrity of the IoT monitoring network data. In addition, detection models can be introduced to detect possible malicious data injection or modification.
\section{Conclusions}
\label{sec:conclusions}
This article has reviewed the use of graph-based models, based on frameworks such as GSP, ML over graphs, or GNNs, to improve data quality in IoT monitoring networks. The basics of the different graph-based frameworks that allow building the necessary models in the context of IoT have been presented. Furthermore, a taxonomy has been proposed to identify different applications that can be performed to improve data quality, such as missing value imputation, anomaly detection, or virtual sensing. All these applications can improve different dimensions of IoT data quality such as completeness, accuracy, or consistency. In addition, this taxonomy makes it possible to identify applications that practitioners can use, depending on their scenario, to meet some data quality requirements. The different applications have been classified according to the scope of the application, i.e., whether the objective is at the network level, at the edge or sensor relationship level, and at the node or sensor level. Once the different applications have been described, different limitations and challenges that may arise when using graph models have been identified. Challenges such as the use of models in large-scale IoT deployments, data availability and heterogeneity, and security. Finally, several emerging trends have been described, such as the development of digital twins using graph-based models or quantum graph models, as well as other areas that seek to provide solutions to the aforementioned limitations, e.g., the generalization and transfer of models, the distributed or federated training, or the security of graph-based models.

\section*{Acknowledgments}

This work is supported by Grant PID2022-138155OB-I00 funded by MCIN/AEI/ 10.13039/501100011033 and by “ERDF A way of making Europe”, CDTI MIG-20221061,  regional project 2021SGR-01059, and by AGAUR 2023 CLIMA 0097.

\bibliographystyle{elsarticle-num}
\bibliography{references}

\end{document}